\documentclass{article}

\PassOptionsToPackage{numbers, compress}{natbib}

\usepackage[eandd, preprint]{neurips_2026}

\usepackage[utf8]{inputenc} 
\usepackage[T1]{fontenc}    
\usepackage{hyperref}       
\usepackage{url}            
\usepackage{booktabs}       
\usepackage{amsfonts}       
\usepackage{nicefrac}       
\usepackage{microtype}      
\usepackage{xcolor}         
\usepackage{graphicx}
\usepackage{multirow}
\usepackage{adjustbox}
\usepackage{xcolor}

\newcommand{\thinkpack}{\textsc{ThinkPack}}
\newcommand{\qwen}{\texttt{Qwen3-8B}}
\newcommand{\qwenbig}{\texttt{Qwen3-14B}}
\newcommand{\olmo}{\texttt{Olmo-3-7B}}
\newcommand{\llama}{\texttt{Llama-R1-8B}}
\newcommand{\nemotron}{\texttt{Nemotron-7B}}

\newcommand{\gsmk}{\texttt{GSM8K}}
\newcommand{\evalplus}{\texttt{EvalPlus}}
\newcommand{\chemistry}{\texttt{Chemistry}}
\newcommand{\orcamath}{\texttt{OrcaMath}}
\newcommand{\selfoss}{\texttt{SelfOSS}}

\newcommand{\vr}{\textit{VR}}
\newcommand{\er}{\textit{ER}}
\newcommand{\mr}{\textit{MR}}
\newcommand{\tr}{\textit{TR}}
\newcommand{\pass}{\textit{pass@1}}
\newcommand{\rpass}{\textit{Rpass@1}}

\newcommand{\masked}[1]{\colorbox{red!35}{#1}}
\newcommand{\open}[1]{\colorbox{green!25}{#1}}

\newcommand{\pypiurl}{\url{https://pypi.org/project/thinkpack/}}
\newcommand{\repourl}{\url{https://github.com/itsluketwist/thinkpack/}}

\title{Reasoning-Trace Collapse: Evaluating the Loss of Explicit Reasoning During Fine-Tuning}

\author{%
  Lukas Twist \\
  King's College London \\
  London, UK \\
  \And
  Helen Yannakoudakis \\
  King's College London \\
  London, UK \\
  \And
  Jie M. Zhang \\
  King's College London \\
  London, UK \\
}

\begin{document}

\maketitle

\begin{abstract}


Explicit reasoning models are trained to produce intermediate reasoning traces before final answers, but downstream fine-tuning is often performed on ordinary instruction--response data that contains no such traces.
We show that this mismatch can induce \emph{reasoning-trace collapse}: a fine-tuned model continues to produce plausible final answers while losing the structurally valid explicit reasoning traces that made it a reasoning model in the first place.
We introduce a structural evaluation framework that separates answer correctness from reasoning-trace validity, measuring valid, empty, missing, and truncated reasoning alongside reasoning-conditioned task performance.
Using this framework, we study four open-weight reasoning models and find that standard supervised fine-tuning can rapidly suppress valid reasoning traces, and that answer-only metrics can substantially obscure this failure: in several settings, performance conditional on valid reasoning remains high while the rate of valid reasoning falls sharply.
We further show that simple loss-masking strategies can substantially mitigate collapse without requiring teacher-generated reasoning traces.
These results suggest that evaluations of fine-tuned reasoning models should report structural reasoning reliability metrics in addition to final-answer performance, especially when adaptation data does not contain explicit reasoning traces.

\end{abstract}

\section{Introduction}\label{sec:intro}

Small language models are increasingly adopted in real-world settings due to their efficiency, low latency, and suitability for deployment under resource constraints~\citep{wangComprehensiveSurveySmall2025}.
In contrast to large, general-purpose models, smaller models are routinely customised via fine-tuning, enabling organisations to adapt them to domain-specific tasks using relatively small instruction--response datasets~\citep{huLoRALowRankAdaptation2021a,zhangInstructionTuningLarge2026}.
Recent advances have extended these models with explicit reasoning capabilities -- where models are trained to produce intermediate reasoning traces before generating a final answer -- to improve performance on tasks requiring multi-step logic~\citep{wangShortSurveySmall2025}.
As a result, small reasoning models are becoming an increasingly practical foundation for real-world systems.

However, current evaluation practice can miss a basic structural failure mode in these models.
Reasoning models are usually evaluated through final-answer performance, such as accuracy or \pass, while the continued presence of explicit reasoning is often treated as implicit.
This is risky because a model may remain capable of producing plausible or correct final answers while losing the behaviour that distinguishes it as a reasoning model: producing a complete, non-empty reasoning trace before its final response.
In this setting, answer-only evaluation can make a fine-tuned reasoning model appear functional even when explicit reasoning has largely disappeared.

This risk is especially important during fine-tuning.
Many datasets used for model customisation do not contain explicit reasoning traces~\citep{guhaOpenThoughtsDataRecipes2025,zhangInstructionTuningLarge2026}, creating a mismatch between reasoning-aware model behaviour and standard downstream adaptation data.
When trained on such data, a model can minimise its loss by learning to produce only the final answer, effectively treating the absence of reasoning as the desired behaviour.
Distillation from a teacher model is one way to bridge this gap~\citep{hsiehDistillingStepbyStepOutperforming2023}, but it requires additional generation and may be impractical when data is private, specialised, or costly to augment.
Lightweight alternatives such as loss masking are widely discussed, but their effectiveness for preserving explicit reasoning traces during fine-tuning remains under-explored~\citep{Qwen3BestPractices,bahreeReasoningAIModels2025}.

We study this problem as an evaluation challenge.
We define \emph{reasoning-trace collapse} as the progressive loss of a model's ability to produce complete, non-empty, structurally valid reasoning traces during fine-tuning.
Unlike work that evaluates the quality, faithfulness, or usefulness of reasoning traces~\citep{leeEvaluatingStepbystepReasoning2025,hao2024llm}, we focus on a more basic question: whether the model produces a valid reasoning trace at all.
This distinction matters because reasoning quality can only be evaluated when explicit reasoning is still present.

To make this failure mode measurable, we introduce a structural evaluation framework that separates answer correctness from reasoning-trace validity.
The framework measures whether generated traces are valid, empty, missing, or truncated, and relates these structural outcomes to task performance through reasoning-conditioned \pass.
This allows us to distinguish cases where a model performs poorly because its reasoning is ineffective from cases where it performs poorly because it no longer produces valid reasoning often enough.

We implement this framework in \thinkpack, a lightweight library for reasoning-aware training, parsing, and evaluation.
Reasoning models use different chat templates, prompt formats, and conventions for representing reasoning traces, so prompt construction, trace extraction, validation, metric computation, and loss masking are often implemented with model-specific code.
\thinkpack\ provides model-agnostic utilities for these operations, allowing the same evaluation and mitigation pipeline to be applied across reasoning formats.


Using this framework, we study \textit{reasoning-trace collapse} in four representative open-weight reasoning models, which differ in both their training pipelines and reasoning formats.
We fine-tune these models on standard instruction--response data without explicit reasoning traces, and evaluate them throughout training on new-task science questions, mathematical reasoning, and code generation.
Our results show that ordinary fine-tuning can rapidly suppress valid reasoning traces, even when final-answer performance is partially preserved.
In several settings, performance conditioned on valid reasoning remains high while the rate of valid reasoning falls sharply, showing that answer accuracy alone can obscure the loss of explicit reasoning behaviour.

We then evaluate mitigation strategies for preserving reasoning traces during fine-tuning.
We compare loss masking against standard fine-tuning and a simple teacher-distillation baseline.
Rather than generating new reasoning traces, masking removes the loss signal from regions where reasoning is absent, preventing the model from being directly rewarded for omitting reasoning.
Across models and reasoning formats, we find that simple masking strategies can substantially preserve valid reasoning behaviour while maintaining strong task performance.
These results suggest that \textit{reasoning-trace collapse} is a practical risk when adapting explicit reasoning models with standard fine-tuning pipelines, but also one that can be measured and mitigated with lightweight interventions.

Our findings highlight a gap in \textit{how} reasoning models are evaluated.
Final-answer metrics remain essential, but they are insufficient for understanding whether explicit reasoning behaviour survives adaptation.
We argue that reasoning-model evaluations should report structural reasoning reliability metrics alongside accuracy, especially after further adaptation.

Our contributions are as follows:

\begin{itemize}
    \item We identify and formalise \emph{reasoning-trace collapse}, a structural failure mode in which explicit reasoning models lose the ability to produce complete, non-empty reasoning traces during fine-tuning, even when task accuracy is preserved.

    \item We introduce a structural evaluation framework that separates answer correctness from reasoning-trace validity, measuring valid reasoning and decomposing invalid reasoning into empty, missing, and truncated traces alongside reasoning-conditioned \pass.

    \item We release \thinkpack, a lightweight library for applying this evaluation framework across reasoning models, providing model-agnostic chat construction, reasoning parsing, loss masking, and metric computation.

    \item We show empirically that standard instruction fine-tuning can rapidly induce reasoning-trace collapse across small reasoning models, and that simple masking strategies substantially preserve explicit reasoning behaviour without requiring teacher-generated reasoning traces.
\end{itemize}


\section{Related Work}

\paragraph{Explicit reasoning in language models.}
Reasoning behaviour in language models has been studied through emergent reasoning in large models~\citep{weiEmergentAbilitiesLarge2022a}, chain-of-thought prompting~\citep{weiChainofthoughtPromptingElicits2022}, and post-training methods that teach models to produce structured reasoning traces~\citep{xuLargeReasoningModels2025}.
Open reasoning models such as \texttt{DeepSeek-R1}~\citep{deepseek-aiDeepSeekR1IncentivizingReasoning2025} have made this behaviour increasingly common in smaller models, typically through combinations of supervised fine-tuning, reinforcement learning, and distillation~\citep{wangShortSurveySmall2025}.
We study what happens after such models are adapted further: whether the explicit reasoning traces they were trained to produce survive downstream fine-tuning.

\paragraph{Evaluating reasoning traces.}
Much work evaluates the quality of reasoning traces, including step-level verification~\citep{lightmanLetsVerifyStep2023a}, fine-grained logical analysis~\citep{zhouDissectingLogicalReasoning2025}, and robustness or consistency across settings~\citep{zhaoComprehensiveEvaluationMultilingual2026}.
These approaches are important, but they generally assume that a reasoning trace is present.
Our work treats structural reasoning \textit{validity} as the central object of evaluation, measuring whether fine-tuned models continue to produce complete reasoning traces and how this interacts with final-answer performance.

\paragraph{Fine-tuning and reasoning preservation.}
Supervised fine-tuning remains the standard way to adapt language models to downstream tasks~\citep{zhangInstructionTuningLarge2026}, but many instruction--response datasets do not contain explicit reasoning traces~\citep{guhaOpenThoughtsDataRecipes2025}.
Prior work shows that fine-tuning can degrade reasoning performance and affect chain-of-thought faithfulness~\citep{loboImpactFineTuningChainofThought2025}.
A common solution is to add reasoning traces through distillation methods, including step-by-step distillation~\citep{hsiehDistillingStepbyStepOutperforming2023}, self-consistent chain-of-thought distillation~\citep{wangSCOTTSelfConsistentChainofThought2023}, and keypoint-based progressive distillation~\citep{fengKeypointbasedProgressiveChainofthought2024}.
These methods are powerful, but require teacher-generated traces or complex pipelines.
In contrast, we study a lighter setting: whether explicit reasoning can be preserved when only ordinary instruction--response data is available.
Loss masking has been suggested for this purpose~\citep{bahreeReasoningAIModels2025,Qwen3BestPractices}, but its effectiveness for preventing the loss of explicit reasoning during fine-tuning remains under-explored.

\paragraph{Tools and frameworks for reasoning models.}
Existing frameworks support language model pipelines and reasoning systems: DSPy optimises LM pipelines~\citep{khattabDSPyCompilingDeclarative2023a}, while LLM Reasoners provides infrastructure for implementing reasoning algorithms~\citep{hao2024llm}.
These tools focus primarily on composing or improving reasoning procedures.
In contrast, much model-level experimentation is built directly on the Hugging Face \texttt{transformers} ecosystem~\citep{TransformersHuggingFace}, which provides flexibility but limited standardisation for reasoning-specific operations across model formats.
\thinkpack\ targets this gap, providing lightweight support for consistent interaction, training, and evaluation of explicit reasoning models.


\section{Reasoning-Trace Collapse: Definition \& Evaluation Framework}
\label{sec:collapse}

Explicit reasoning models are trained to produce intermediate reasoning traces before their final answers, using structured delimiters such as \texttt{<think>}...\texttt{</think>}.
This behaviour is typically introduced through specialised post-training pipelines, including reinforcement learning, distillation, and format-based supervision~\citep{tieSurveyPosttrainingLarge2025}.
While these pipelines can successfully induce explicit reasoning, behaviours introduced during post-training may be fragile under further model customisation~\citep{qiFinetuningAlignedLanguage2023}.

This becomes a practical problem when reasoning models are adapted to new instruction--response data, many of which provide target answers without explicit reasoning traces~\citep{guhaOpenThoughtsDataRecipes2025,zhangInstructionTuningLarge2026}.
Fine-tuning can therefore teach the model to produce the final answer, but it provides no signal to preserve the explicit reasoning behaviour introduced during post-training.
This creates the setting in which \textit{reasoning-trace collapse} can occur.

\paragraph{Definition.}
We define \emph{reasoning-trace collapse} as the progressive loss of a model's ability to produce structurally valid reasoning traces during fine-tuning.

Formally, given an input $x$, a reasoning model produces an output $y=(r,a)$, where $r$ is a reasoning trace and $a$ is the final answer.
Reasoning-trace collapse occurs when $r$ is no longer reliably produced or is not structurally valid, even if $a$ remains correct.

This definition is deliberately structural.
It does not claim that a valid trace is faithful, correct, or useful; it only captures whether explicit reasoning is present in a form that can be parsed and evaluated.
This distinguishes \textit{reasoning-trace collapse} from broader reasoning degradation during fine-tuning~\citep{loboImpactFineTuningChainofThought2025}: a model may perform worse because its reasoning quality declines, or because it stops producing valid reasoning traces often enough.
Final-answer metrics alone cannot separate these cases.

\paragraph{Structural reasoning-trace validity.}
We argue that evaluations of explicit reasoning models should track whether generated reasoning traces are structurally valid, not only whether the final answer is correct.
A response has \emph{valid reasoning} when its reasoning trace is complete, non-empty, and can be reliably separated from the final answer.
Invalid reasoning is decomposed into three failure types: \emph{empty}, where reasoning delimiters are present but contain no content; \emph{missing}, where no reasoning trace can be extracted; and \emph{truncated}, where a reasoning trace begins but cannot be parsed as complete -- this includes cases where generation ends before the reasoning delimiter is closed, or the output reaches the generation limit before completing the expected reasoning structure.
We report valid reasoning rate (\vr) as the main structural reliability metric, and use empty reasoning rate (\er), missing reasoning rate (\mr), and truncated reasoning rate (\tr) to explain how invalid reasoning occurs.
These metrics are deliberately structural: they do not judge whether a trace is correct or faithful, but whether explicit reasoning is present in a form that can be parsed, measured, and evaluated.

\paragraph{Reasoning-conditioned performance.}
Structural reasoning-trace validity should also be related back to task success.
For this, we introduce reasoning-conditioned accuracy (\rpass): \pass\ computed only over generations that contain valid reasoning traces.
While \pass\ measures overall task performance, \rpass\ measures performance conditional on the model successfully producing structurally valid reasoning.
This distinction helps separate different failure modes.
If both \vr\ and \rpass\ fall, the model is reasoning less often and performing worse when it does reason.
If \vr\ falls while \rpass\ remains stable or improves, the model's remaining valid reasoning is still effective, but the model is becoming less likely to produce it.
We treat this second pattern as the characteristic signature of \textit{reasoning-trace collapse}.

\paragraph{A structural evaluation framework.}
Reasoning-trace collapse exposes a blind spot in answer-only evaluation.
Final-answer metrics remain necessary, but they are insufficient for evaluating explicit reasoning models after adaptation.
A fine-tuned model may appear functional, or even improve on an in-domain benchmark, while losing the structurally valid reasoning traces that make reasoning-specific evaluation possible.
Our framework addresses this by separating answer correctness from reasoning-trace validity: \vr, \er, \mr, and \tr\ measure whether explicit reasoning is structurally present, while \rpass\ measures task performance conditioned on valid reasoning.
Together, these metrics help distinguish cases where performance drops because reasoning quality declines from cases where the model simply stops producing valid reasoning often enough.
The models may not be getting worse at these tasks; they may simply be forgetting to produce explicit reasoning before answering.
For this reason, evaluations of fine-tuned reasoning models should report structural reasoning reliability alongside final-answer performance, particularly when adaptation data does not contain explicit reasoning traces.


\section{\thinkpack: Operationalising Structural Reasoning Evaluation}
\label{sec:thinkpack}

Measuring reasoning-trace collapse requires consistent prompt construction, trace parsing, structural metric computation, and loss masking across model formats.
These steps are not standardised: reasoning models differ in chat templates, delimiters, and conventions for representing intermediate traces, so experimental pipelines are often reimplemented with model-specific code.

We introduce \thinkpack, a lightweight framework for reasoning-aware training and evaluation.
The goal of \thinkpack\ is not to introduce a new reasoning algorithm or training method, but to provide a common interface for applying our structural evaluation framework across models.
It integrates directly with the Hugging Face \texttt{transformers} ecosystem~\citep{TransformersHuggingFace} and exposes model-agnostic utilities for chat construction, trace parsing, metric computation, and loss masking.
This allows the same evaluation and mitigation pipeline to be applied across models with different reasoning conventions, reducing implementation-specific code and making reasoning-aware analysis more reproducible.

\paragraph{Universal reasoning interface.}
\thinkpack\ provides a reasoning-aware wrapper around \texttt{tokenizer.apply\_chat\_template}, mapping explicit reasoning fields to the appropriate model format automatically.
The same interface supports models that generate reasoning directly in text, use separate reasoning fields, or rely on prefixed reasoning templates.
It can therefore be used both for evaluation-time prompt construction and training-time data formatting.
\thinkpack\ also exposes model-agnostic loss-masking utilities, allowing prompts, missing reasoning regions, and final responses to be included or excluded from the training loss in a controlled way.

\paragraph{Trace parsing and reasoning statistics.}
\thinkpack\ parses generated outputs into structured reasoning and answer segments, then classifies reasoning as valid, empty, missing, or truncated according to the framework defined in Section~\ref{sec:collapse}.
Using these parsed outputs, it computes the reasoning-reliability metrics used throughout this paper, including \vr, \er, \mr, \tr, and \rpass.
This provides a consistent basis for comparing reasoning-trace validity across models, rather than relying on ad hoc string matching or model-specific evaluation code.

\paragraph{Availability.}
\thinkpack\ is released as an open-source Python package and is available on both PyPI\footnote{\pypiurl} and GitHub.\footnote{\repourl}
The repository includes documentation, examples, tests, and contribution guidelines.
\textit{Further artefact details and minimal code examples are provided in Appendix~\ref{app:thinkpack}.}


\section{Empirical Study}
\label{sec:empirical}

We use \thinkpack\ to study reasoning-trace collapse during supervised fine-tuning of small reasoning models.
Our experiments address two questions.
First, how does the representation of missing reasoning in the fine-tuning data affect reasoning-trace collapse?
Second, can reasoning-aware loss masking preserve explicit reasoning, and how does it compare with teacher-generated reasoning traces?

This design reflects the practical setting that motivates our work.
Reasoning models are often adapted using instruction--response datasets that contain target answers but no explicit reasoning traces~\citep{zhangInstructionTuningLarge2026}.
When such data is formatted for training, missing reasoning must still be represented somehow, either by including an empty reasoning block or by omitting reasoning tags entirely.
Our experiments test whether reasoning-trace collapse can be explained and reduced through lightweight changes to the fine-tuning pipeline.


\subsection{Experimental Set-up}\label{sec:setup}


\paragraph{Models.}
We study four representative small reasoning models that differ in their training pipelines and reasoning formats.
\qwen~\citep{QwenQwen38BHugging2025} uses a chat template with an explicit reasoning field and expects the model to produce the full reasoning block during generation.
\olmo~\citep{AllenaiOlmo37BThinkHugging2026}, in contrast, uses a prefixed reasoning format in which the opening reasoning tag is inserted by the template, encouraging reasoning to begin in every response.
\llama~\citep{deepseek-aiDeepSeekR1IncentivizingReasoning2025} is a \texttt{Llama-8B} model adapted primarily through distillation from \texttt{DeepSeek-R1}-generated reasoning data.
\nemotron~\citep{NvidiaOpenReasoningNemotron7BHugging2026} is part of NVIDIA's OpenReasoning line of models, and is a Qwen2.5-7B derivative post-trained with supervised reasoning data.
This broadens our coverage across reasoning formats, base architectures, and post-training pipelines, while keeping the study focused on controlled comparisons.

\paragraph{Training Dataset.}
We fine-tune on science question answering, a domain distinct from the mathematics and code tasks commonly used to evaluate reasoning models.
Specifically, we use the Chemistry L-3 subset of \texttt{SciKnowEval}~\citep{fengSciKnowEvalComprehensiveDataset2025}, augmented with answer explanations from prior work~\citep{shenfeldSelfDistillationEnablesContinual2026}.
Each response is formatted as an explanation followed by the final answer, but does not include explicit reasoning delimiters or a separate reasoning trace.
This gives us a realistic instruction--response setting: the target responses contain useful explanatory content, but are not formatted as reasoning-model outputs.

\paragraph{Training Configuration.}
All models are fine-tuned for three epochs using non-quantised LoRA~\citep{huLoRALowRankAdaptation2021a}, AdamW optimisation~\citep{loshchilovDecoupledWeightDecay2018} with an initial learning rate of $1\mathrm{e}{-5}$, and a cosine learning rate schedule with warm-up.
Results in the main paper are averaged over two random seeds.
All models are trained and evaluated on NVIDIA H200 GPUs.
\textit{Further training details are provided in Appendix~\ref{app:experimental_details}.}
\textit{We selected a stable learning rate from an initial sweep and report the full sensitivity analysis in Appendix~\ref{app:lr_sweep}.}

\paragraph{Evaluation.}
We evaluate both in-domain adaptation and out-of-domain generalisation.
For in-domain performance, we use the held-out evaluation split of the Chemistry L-3 subset of \texttt{SciKnowEval} (\chemistry).
For out-of-domain performance, we evaluate mathematical reasoning with \gsmk~\citep{cobbeTrainingVerifiersSolve2021b} and code generation with \evalplus~\citep{liuYourCodeGenerated2023a}.
To track reasoning behaviour throughout fine-tuning, we evaluate every 100 optimisation steps, giving 20 evaluations over the three-epoch training run.
For efficiency, each evaluation uses a fixed 256-sample subset from each dataset, so results should be interpreted as controlled within-study comparisons rather than full-benchmark estimates.
All evaluations use greedy decoding and no system prompt, isolating the effect of fine-tuning from prompt engineering or sampling variation.

\paragraph{Metrics.}
We report standard task accuracy using \pass, alongside reasoning-reliability metrics computed with \thinkpack.
Our primary structural metric is \textit{valid reasoning rate (\vr)}, which measures how often the model produces a complete, non-empty reasoning trace that can be reliably separated from the final answer.
We decompose invalid reasoning into \textit{empty}, \textit{missing}, and \textit{truncated} reasoning, and report the corresponding rates (\er, \mr, and \tr) to show how reasoning fails.
We also report reasoning-conditioned \pass\ (\rpass): accuracy computed only over responses with valid reasoning, suppressing it when there are 10 or fewer valid-reasoning responses because the estimate becomes unstable on very small subsets.
Together, these metrics allow us to distinguish between models that fail because their valid reasoning performs poorly, and models that fail because they stop producing valid reasoning at all.


\paragraph{Ablations.}
We focus the main experiments on four small reasoning models fine-tuned on the \chemistry\ dataset.
To test whether the findings are sensitive to model scale or the choice of fine-tuning dataset, we include additional ablations with \qwenbig, and with \qwen\ fine-tuned on math and code datasets, in Appendix~\ref{app:training_ablations}.
Further analyses of training-seed stability, evaluation-subset robustness, reasoning quality, and truncation mechanisms are provided in Appendix~\ref{app:additional-analyses}.


\subsection{Experiment 1: Format-Induced Reasoning-Trace Collapse}
\label{sec:experient1}

\begin{figure*}
    \centering
    \includegraphics[width=\textwidth]{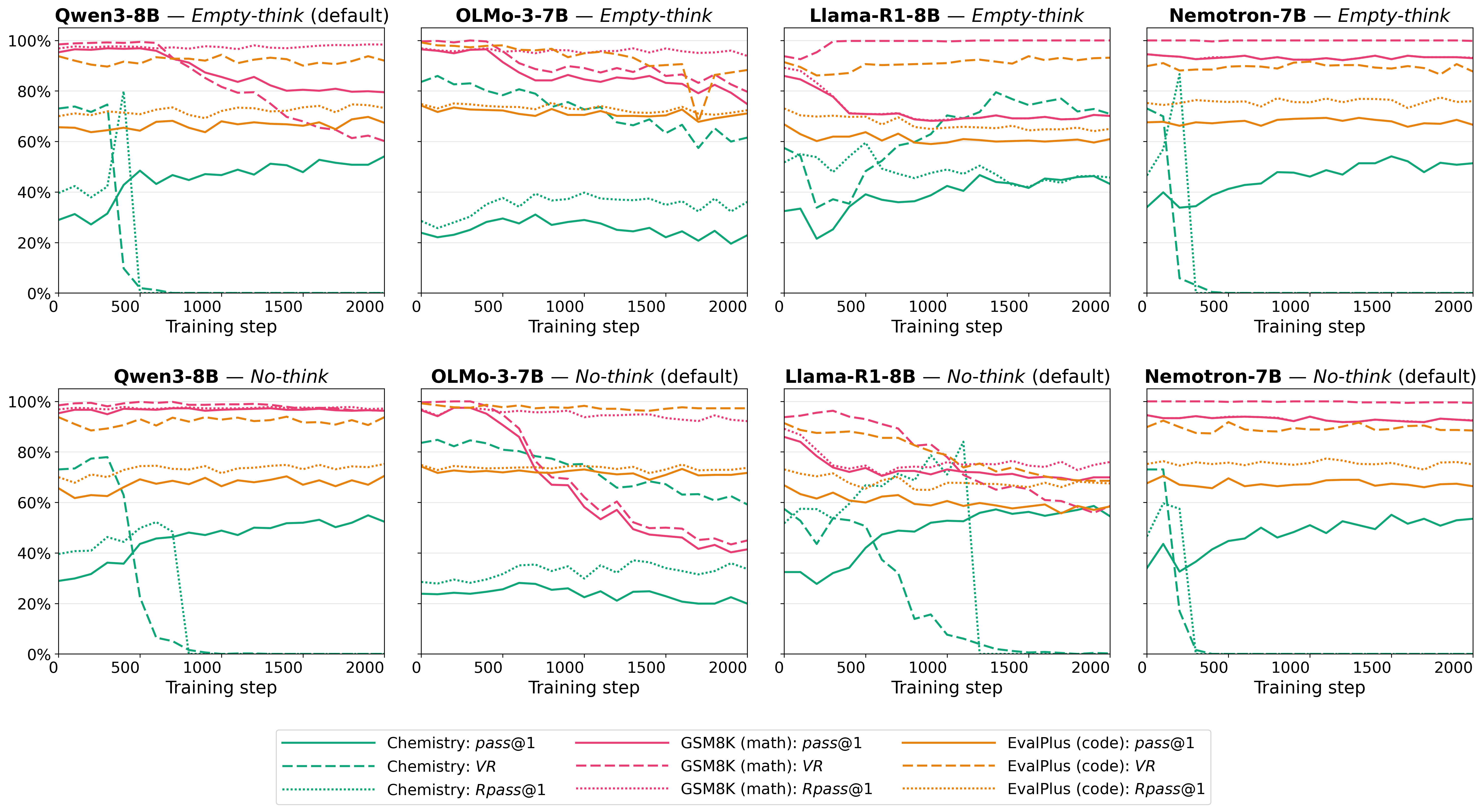}
    \caption{
        \textbf{\textit{Format-Induced Reasoning-Trace Collapse.}}
        We compare two ways of representing missing reasoning during supervised fine-tuning: including an empty \texttt{<think>} block, or omitting reasoning tags entirely.
        Metrics are measured every 100 training steps across three datasets: \chemistry, \gsmk\ (math), and \evalplus\ (code).
        Solid lines show \pass, dashed lines show valid reasoning rate (\vr), and dotted lines show reasoning-conditioned \pass\ (\rpass).
        Metrics are averaged over two random seeds.
    }
    \label{fig:format}
\end{figure*}

\paragraph{Set-up.}
We first test whether reasoning-trace collapse depends on how missing reasoning is represented during fine-tuning.
We compare two formats for instruction--response data without explicit reasoning traces: \textit{empty-think}, which inserts an empty reasoning block before the final answer, and \textit{no-think}, which omits the reasoning block entirely.
These correspond to different model defaults: \qwen\ defaults to \textit{empty-think}, while \olmo, \llama\ and \nemotron\ default to \textit{no-think}.
By evaluating both formats for all models, we test whether the expected model convention is actually the most protective against reasoning-trace collapse.


\paragraph{Results.}
Figure~\ref{fig:format} shows that the representation of missing reasoning has a large effect on reasoning-trace collapse across models.
The strongest effect appears on the in-domain \chemistry\ task.
For \qwen\ and \nemotron, valid reasoning collapses in both formats, reaching 0\% final \vr, while \pass\ continues to improve to 50--57\%.
For \llama, the effect is strongly format-dependent: the default \textit{no-think} setting collapses \chemistry\ reasoning to 0\% \vr, while \textit{empty-think} preserves 71\% \vr.
\olmo\ shows a different profile again, with more gradual degradation and final \chemistry\ \vr\ around 59--62\%, driven primarily by truncated reasoning.
These results show why answer-only metrics are insufficient: models can continue adapting to the new task while losing the structurally valid reasoning traces needed for reasoning-aware evaluation.

The effect is not limited to the fine-tuning domain.
On \gsmk, \qwen\ loses valid reasoning under its default \textit{empty-think} format, falling to 60\% \vr\ and 80\% \pass, while \rpass\ remains high at 98\%.
This is consistent with reduced trace validity contributing to the performance decline, while accuracy among the remaining valid-reasoning responses stays high.
\olmo\ shows the same pattern in both settings, but collapses much more sharply under its default \textit{no-think} format, where final \vr\ falls to 45\% compared with 80\% for \textit{empty-think}.
\llama\ follows a similar format-dependent pattern: \textit{no-think} leaves only 58\% \gsmk\ \vr, while \textit{empty-think} preserves 100\% \vr.
In contrast, \nemotron\ remains stable on \gsmk\ in both formats, preserving approximately 100\% \vr.
Across models, \evalplus\ is generally less sensitive, although \llama\ again shows degradation under \textit{no-think}, falling to 69\% \vr\ compared with 93\% under \textit{empty-think}.

Overall, the most notable result is that model-default formatting is not reliably protective.
\qwen\ is more stable when reasoning tags are omitted, while \olmo\ and \llama\ preserve reasoning more effectively when empty reasoning tags are included; for \nemotron, behaviour is similar across formats.
This shows that the representation of missing reasoning can directly shape whether explicit reasoning survives fine-tuning, and that default formatting should be treated as part of the experimental setup rather than assumed to be protective.
The breakdown of invalid reasoning is also essential: \qwen, \llama, and \nemotron\ collapse mainly through empty or missing traces, while \olmo\ collapses mainly through truncation.
Reporting \vr\ tells us whether valid reasoning has been lost; reporting \er, \mr, and \tr\ tells us how.
\textit{The full metrics breakdown is available in Appendix~\ref{app:full_metrics}.}

\begin{figure*}
    \centering
    \includegraphics[width=\textwidth]{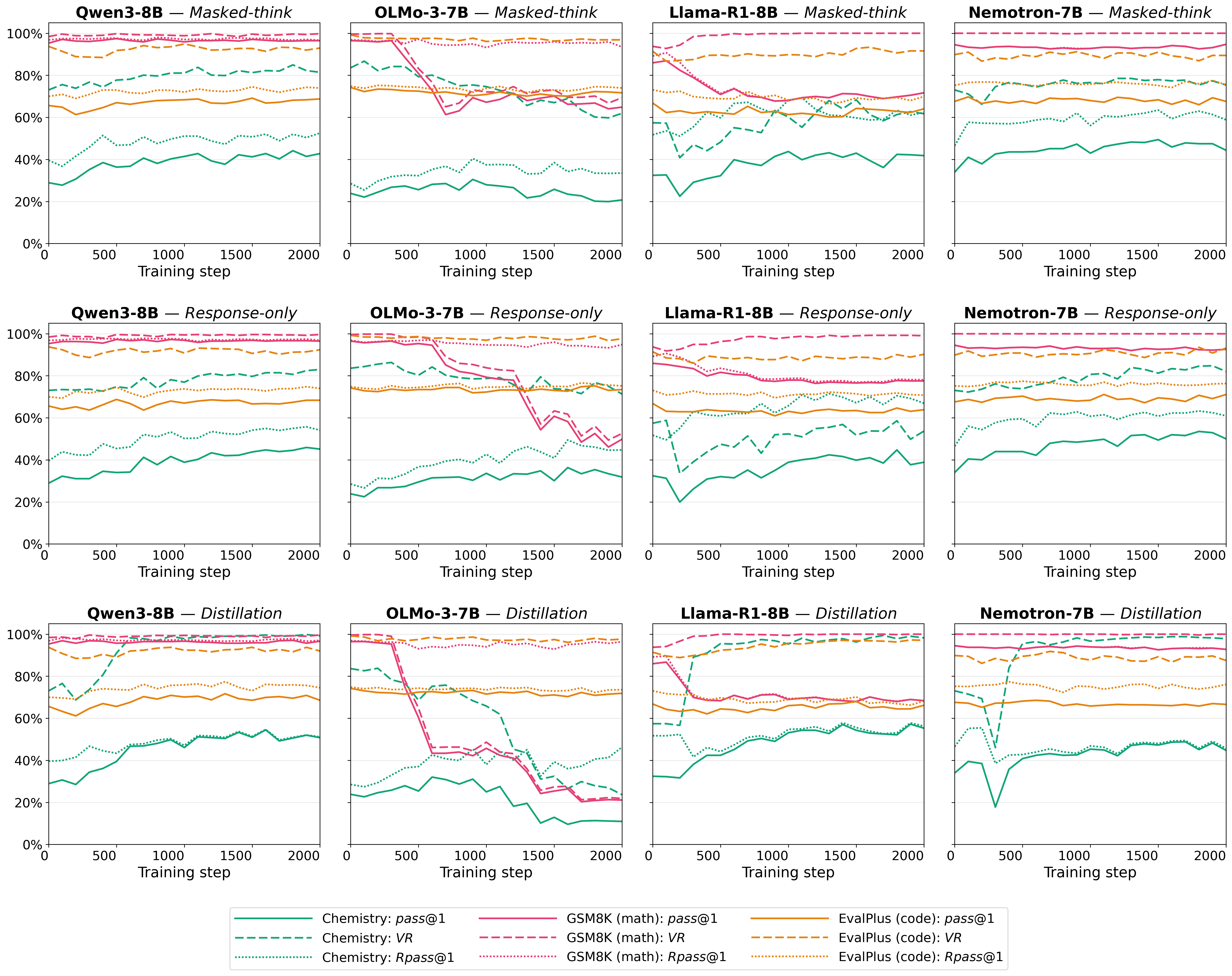}
    \caption{
        \textbf{\textit{Mitigating Reasoning-Trace Collapse.}}
        We compare three ways to mitigate reasoning-trace collapse: masking the empty \texttt{<think>} block, updating weights using the response \textit{only}, or using distillation from a teacher model.
        Metrics are measured every 100 training steps across three datasets: \chemistry, \gsmk\ (math), and \evalplus\ (code).
        Solid lines show \pass, dashed lines show valid reasoning rate (\vr), and dotted lines show reasoning-conditioned \pass\ (\rpass).
        Metrics are averaged over two random seeds.
    }
    \label{fig:masking}
\end{figure*}


\subsection{Experiment 2: Mitigating Reasoning-Trace Collapse}
\label{sec:experient2}

\paragraph{Set-up.}
We next evaluate whether reasoning-trace collapse can be mitigated during fine-tuning.
We compare two lightweight loss-masking strategies against a teacher-distillation baseline.
For \textit{masked-think} and \textit{response-only}, the training data is formatted with empty \texttt{<think>} tags.
In the \textit{masked-think} setting, the empty reasoning region is excluded from the loss, preventing the model from being directly rewarded for producing an empty trace.
In the \textit{response-only} setting, both the prompt and reasoning are masked, so only the final response contributes to the training loss.
For \textit{distillation}, we generate reasoning traces for the training data using \texttt{GPT-5-mini}~\citep{openaiGPT5MiniAPI2025}, and fine-tune on the resulting reasoning-augmented dataset, providing a teacher-augmented but more expensive baseline (Appendix~\ref{app:distillation_details}).


\paragraph{Results.}
Figure~\ref{fig:masking} shows that mitigation effectiveness is strongly model-dependent.
For \qwen\ and \nemotron, all three mitigation strategies preserve high \vr\ on \gsmk\ and \evalplus, with little change in overall task performance.
On the in-domain \chemistry\ task, both masking strategies allow adaptation while retaining much more valid reasoning than standard fine-tuning.
Distillation is especially strong for preserving reasoning structure, giving the highest observed \chemistry\ \vr\ for both models.
For \qwen, peak \chemistry\ \pass\ remains modest under masking (44--47\%), compared with 55\% under distillation.
For \nemotron, the pattern is slightly different: masking gives the stronger task performance, peaking at 56\% for \textit{response-only} and 51\% for \textit{masked-think}, compared with 49\% for distillation.
In both models, masking also changes the failure profile: rather than collapsing into empty or missing reasoning, most remaining invalid traces are truncated.

For \llama, mitigation is also effective, but the trade-offs differ by strategy.
Both masking strategies avoid the complete \chemistry\ reasoning collapse seen under standard fine-tuning, though final \chemistry\ \vr\ reaches only 62\% for \textit{masked-think} and 54\% for \textit{response-only}.
The out-of-domain results show a similar split.
\evalplus\ remains relatively stable across strategies, but while \gsmk\ \vr\ improves substantially under masking it sees drops in both \pass\ and \rpass.
Distillation is especially effective for preserving \llama's reasoning structure, reaching near 100\% \vr\ for all benchmarks.
However, \pass\ and \vr\ still move differently across tasks, reinforcing the need to track both final-answer performance and structural reasoning reliability.

\olmo\ remains the most difficult case.
Unlike the other models, it shows limited \chemistry\ adaptation across mitigation strategies, and its reasoning failures are dominated by truncation.
On \evalplus, all strategies remain relatively robust, but on \gsmk\ none fully prevents reasoning-trace collapse.
\textit{Masked-think} preserves more final valid reasoning than \textit{response-only} (69\% vs. 53\%), although both show sharp drops during training.
Most notably, distillation performs worst for \olmo: despite training on teacher-generated traces, final \gsmk\ \vr\ falls to 22\%, and \chemistry\ \vr\ falls to 24\%.
This suggests that neither masking nor teacher traces fully resolve \olmo's tendency to produce incomplete reasoning.
Further analysis shows that most \gsmk\ truncation occurs before the generation budget is exhausted (Appendix~\ref{app:olmo-analysis}).

Overall, masking shows that reasoning-trace collapse is not inevitable when fine-tuning on data without explicit reasoning traces.
Loss masking preserves much of the model's structurally valid reasoning while still allowing adaptation to the new instruction--response data, making it a practical alternative when teacher-generated traces are unavailable, expensive, or unreliable.
At the same time, the distillation results are a useful warning: the same basic teacher-trace pipeline is highly effective for \qwen\ and \llama, and preserves strong reasoning for \nemotron, but performs poorly for \olmo.
Mitigation is therefore both strategy- and model-dependent, and should be evaluated with structural reasoning metrics rather than assumed from final-answer performance alone.


\section{Discussion}
\label{sec:discussion}


\paragraph{Structural reasoning reliability should be reported alongside task performance.}
Final-answer metrics can hide reasoning-trace collapse.
For \qwen, \llama, and \nemotron, \chemistry\ performance continues to improve even after valid reasoning disappears.
On \gsmk, \pass\ falls with \vr\ for both \qwen\ and \olmo, while \rpass\ remains high.
This shows that models are not necessarily becoming worse when they reason; rather, they can become less likely to produce valid reasoning at all.
For explicit reasoning models, this is a structural failure mode that answer-only evaluation can easily miss.
Tracking reasoning structure could help identify these failures before fine-tuned models are used in downstream applications.
However, treating a structurally valid trace as evidence of correct or trustworthy reasoning could create false confidence in their reliability.

Additional ablations support this pattern beyond the main setting: \qwenbig\ also exhibits reasoning-trace collapse under standard fine-tuning (Appendix~\ref{app:larger_model}, and \qwen\ fine-tuned on math or code shows that collapse is not specific to \chemistry, often appearing most strongly in the domain targeted by fine-tuning (Appendix~\ref{app:other_datasets}).
A small model-assisted audit also finds that the surviving valid traces retain similar judged correctness before and after fine-tuning (Appendix~\ref{app:trace-quality}).
However, this does not establish that the traces causally contribute to the final answers.

\paragraph{Fine-tuning format is part of the evaluation setting.}
Missing-reasoning format is not a minor implementation detail.
Model defaults are not reliably protective: \qwen\ is more stable when reasoning tags are omitted, while \olmo\ and \llama\ preserve reasoning better when empty reasoning tags are included; \nemotron\ behaves similarly across formats.
The failure modes also differ, with \qwen, \llama, and \nemotron\ collapsing mainly through empty or missing reasoning, while \olmo\ mainly collapses through truncation.
Training format should therefore be reported alongside the model, dataset, and optimisation configuration.

\paragraph{Masking is useful, but model-specific.}
Loss masking can preserve valid reasoning without teacher-generated traces, but its effects remain model- and task-dependent.
For \qwen, \llama, and \nemotron, masking maintains high \vr\ on \gsmk\ and \evalplus, while also preserving substantially more valid reasoning on the learned \chemistry\ task.
For \olmo, masking helps in some settings but does not remove its tendency to produce truncated reasoning.
Importantly, basic distillation is not uniformly better: it is highly effective for \qwen\ and \llama, maintains performance for \nemotron\ without clearly improving new-task learning, and still fails to prevent \gsmk\ collapse for \olmo.
This suggests that teacher-generated traces are not an easy or guaranteed solution, and that mitigation strategies still need to be evaluated per model rather than assumed to transfer.
Masking is therefore a practical mitigation, not a complete solution: it preserves the structural presence of reasoning while using the original instruction--response data, but does not guarantee that reasoning is correct, faithful, or useful.


\section{Limitations}\label{sec:limitations}

\paragraph{Tooling Scope.}
\thinkpack\ relies on model tokenizers and chat templates to infer reasoning structure, which means its parser and masking utilities may need updates for new architectures or unusual formatting conventions.
We mitigate this by allowing reasoning formats to be overridden, keeping the parsing logic extensible, and validating the parser through unit tests and manual checks of generated outputs (Appendix~\ref{app:validation}).
\thinkpack\ is also designed primarily for Hugging Face \texttt{transformers}-based workflows.
This makes it easy to integrate with standard model-level experiments, but means higher-level orchestration or serving frameworks may require additional adapters.
Extending support beyond \texttt{transformers} is future work.

\paragraph{Empirical Study External Validity.}
Our main experiments cover four small reasoning models and one fine-tuning dataset.
This controlled design lets us compare distinct reasoning formats and fine-tuning strategies, but does not establish how common reasoning-trace collapse is across all model families, scales, domains, or training regimes.
We mitigate this by selecting models with different reasoning conventions, evaluating both in-domain and out-of-domain tasks, and including targeted appendix ablations over model scale and fine-tuning dataset (Appendix~\ref{app:training_ablations}).
Broader validation across additional model families, datasets, and adaptation settings remains important future work.

\paragraph{Empirical Study Internal Validity.}
Our study prioritises controlled comparison between fine-tuning strategies, but necessarily fixes several design choices.
We use LoRA as a widely used adaptation method for low-data settings~\citep{zhangWhenScalingMeets2024} that can help preserve base model capabilities~\citep{biderman2024lora}.
We also use greedy decoding and fixed 256-example evaluation subsets to reduce variance from sampling and dataset scale.
The uncertainty intervals in Appendix~\ref{app:full_metrics} account for evaluation-example sampling under a conservative dependence assumption, but should not be interpreted as a full estimate of training-run variability.
Reasoning-trace collapse may therefore differ under stochastic decoding, different seeds, larger evaluation sets, or alternative fine-tuning methods.
Finally, our metrics focus on the presence and structure of explicit reasoning, not its correctness, faithfulness, or usefulness; they are intended to complement, not replace, reasoning-quality evaluations.


\section{Conclusion}
\label{sec:conclusion}

We introduced \textit{reasoning-trace collapse}, a structural failure mode in which explicit reasoning models lose the ability to produce complete, non-empty reasoning traces during fine-tuning.
We argued that this failure is largely invisible to final-answer metrics, and introduced a structural evaluation framework that separates answer correctness from reasoning-trace validity.
Using \thinkpack, we showed that ordinary fine-tuning on data without explicit reasoning traces can rapidly induce reasoning-trace collapse, that the representation of missing reasoning strongly affects the severity and form of collapse, and that simple loss-masking strategies can substantially preserve explicit reasoning behaviour.
We also found that teacher distillation is not uniformly stronger than masking, suggesting that lightweight changes to the training objective can be a practical alternative when generated traces are costly or unreliable.
Fine-tuned reasoning models should therefore be evaluated not only by whether they answer correctly, but by whether they still produce the explicit reasoning traces they were designed to provide.

\section*{Acknowledgements \& Disclosures}
This work is supported by the UKRI Centre for Doctoral Training in Safe and Trusted AI (EPSRC iCASE Award ref EP/Y528572/1).
The authors declare no competing interests.

\newpage



\bibliographystyle{plainnat}
\bibliography{main}


\appendix




\newpage

\section{\thinkpack\ Artefact \& Usage Details}
\label{app:thinkpack}

\thinkpack\ is released as a lightweight Python package for training, parsing, and evaluating explicit reasoning models.
It is designed to make reasoning-aware experiments easier to reproduce by standardising the parts of the pipeline that are otherwise often implemented separately for each model: chat construction, response parsing, reasoning-reliability statistics, and loss masking.
This appendix summarises the artefact details and provides minimal code examples of the main functionality used in this paper.

\subsection{Artefact Details}

\thinkpack\ is publicly available on GitHub\footnote{\repourl} and PyPI\footnote{\pypiurl} under the MIT license.
It requires Python 3.11 or later and can be installed with:

\begin{verbatim}
pip install thinkpack
\end{verbatim}

The repository includes source code, examples, tests, development documentation, and contribution guidelines.
It is designed to integrate with standard Hugging Face \texttt{transformers}-based workflows, while handling reasoning-specific formatting and parsing internally.

The package exposes four main modules.
\texttt{thinkpack.chat} applies model-aware chat templates with reasoning-history support and optional thought steering.
\texttt{thinkpack.parse} separates raw model outputs into reasoning and answer components, with flags for missing, empty, truncated, and valid reasoning.
\texttt{thinkpack.stats} aggregates parsed outputs into structural reasoning metrics such as valid reasoning rate and missing reasoning rate.
\texttt{thinkpack.mask} constructs masked fine-tuning datasets, allowing reasoning regions or prompts to be excluded from the training loss.

The repository also includes an \texttt{llms.txt} file and a small command-line utility for installing ThinkPack as an agent skill in tools such as Claude Code, Cursor, and Windsurf.

\subsection{Model-format support}

\thinkpack\ uses the model tokenizer and chat template to identify how reasoning is represented.
Its \texttt{detect\_model(...)} function supports inline reasoning, prefixed reasoning, and reasoning stored in a separate field.
For models whose format cannot be detected automatically, users can provide a custom \texttt{ModelInfo} specifying the reasoning style, delimiters, and reasoning field. 
This definition is shared across the chat, parsing, and masking utilities without requiring changes to the core pipeline.

\thinkpack\ currently targets Hugging Face-compatible tokenizer and chat-template workflows.
Serving libraries that retain these components can also be supported, while provider-specific response schemas or non-Hugging Face inference paths may require additional adapters.

\subsection{Code Examples}

\paragraph{Chat construction.}
The \texttt{chat} module provides a model-aware wrapper around \texttt{tokenizer.apply\_chat\_template}.
It can automatically apply the correct template, embed the reasoning history in the assistant messages, and optionally seed the reasoning or final response:

\begin{verbatim}
import thinkpack

prompt = thinkpack.apply_chat_template(
    conversation=conversation,
    tokenizer=tokenizer,
)

prompt = thinkpack.apply_chat_template(
    conversation=conversation,
    tokenizer=tokenizer,
    think_prefix="Let me break this down step by step.",
    response_prefix="The answer is",
)
\end{verbatim}

Reasoning can also be included directly in assistant messages for multi-turn conversations:

\begin{verbatim}
conversation = [
    {"role": "user", "content": "What is 2 + 2?"},
    {"role": "assistant", "reasoning": "2 + 2 = 4", "content": "4"},
    {"role": "user", "content": "And 3 + 3?"},
]

prompt = thinkpack.apply_chat_template(
    conversation=conversation,
    tokenizer=tokenizer,
)
\end{verbatim}

\paragraph{Parsing reasoning traces.}
The \texttt{parse} module converts raw generations into structured reasoning and answer fields.
It also records the reasoning-validity flags used to measure reasoning-trace collapse:

\begin{verbatim}
parsed = thinkpack.parse(response=raw_text, tokenizer=tokenizer)

parsed.reasoning
parsed.answer
parsed.has_valid_reasoning
parsed.has_missing_reasoning
parsed.has_empty_reasoning
parsed.has_truncated_reasoning
\end{verbatim}

The same function can be applied to a batch of responses:

\begin{verbatim}
parsed_outputs = thinkpack.parse(
    response=responses,
    tokenizer=tokenizer,
)
\end{verbatim}

\paragraph{Computing reasoning statistics.}
The \texttt{stats} module aggregates parsed outputs into reasoning-reliability metrics:

\begin{verbatim}
parsed_outputs = thinkpack.parse(
    response=responses,
    tokenizer=tokenizer,
)

stats = thinkpack.compute_stats(responses=parsed_outputs)

stats.valid_reasoning_rate
stats.missing_reasoning_rate
stats.empty_reasoning_rate
stats.truncated_reasoning_rate
stats.answer_rate
\end{verbatim}

These statistics provide the structural metrics used throughout the paper.
Task-level metrics such as \pass\ and \rpass\ are then computed alongside these reasoning-reliability statistics if the \texttt{results} parameter is used.

\paragraph{Loss masking.}
The \texttt{mask} module formats training records into a pre-tokenised Hugging Face dataset while excluding selected regions from the training loss.
Masking the reasoning block prevents the model from being rewarded for producing empty reasoning during supervised fine-tuning:

\begin{verbatim}
dataset = thinkpack.apply_mask(
    conversations=conversations,
    tokenizer=tokenizer,
    masked=thinkpack.MaskType.THINK,
)
\end{verbatim}

Response-only training is implemented by composing mask flags:

\begin{verbatim}
dataset = thinkpack.apply_mask(
    conversations=conversations,
    tokenizer=tokenizer,
    masked=thinkpack.MaskType.PROMPT | thinkpack.MaskType.THINK,
)
\end{verbatim}

\paragraph{Custom Model Formats.}
For models whose reasoning format is not detected automatically, THINKPACK allows users to provide a custom \texttt{ModelInfo} specifying the reasoning delimiters and formatting conventions.
The following example demonstrates a model that uses \texttt{[REASONING]...[/REASONING]} tags, where the opening tag is already provided by the prompt.

\begin{verbatim}
from thinkpack import ModelInfo, TagStyle, parse

info = ModelInfo(
    prefixed=True,
    tag_content="REASONING",
    tag_style=TagStyle.BRACKET,
)

result = parse(
    response="2 + 2 = 4.[/REASONING]The answer is 4.",
    model_info=info,
)

result.reasoning            # "2 + 2 = 4."
result.answer               # "The answer is 4."
result.has_valid_reasoning  # True
\end{verbatim}

\subsection{Parser Validation}
\label{app:validation}
To validate the parser under the formats used in this study, we manually inspected 100 samples from the generated outputs across models, evaluation tasks, training formats, and checkpoints.
We included examples from all observed parser categories (valid, empty, missing and truncated).
For each output, we labelled the reasoning status as valid, empty, truncated, or missing, and compared this annotation with \thinkpack's parser output.
Agreement was 100\%.
This suggests that the structural categories are reliably captured under the models used in this study.


\section{Experimental Details}
\label{app:experimental_details}

This appendix provides the implementation details omitted from the main paper.
We first describe the shared fine-tuning configuration;
then show how training examples are formatted under the missing-reasoning conditions used in Experiment~1 (Section~\ref{sec:experient1});
finally, we illustrate how loss masks are applied in Experiment~2 (Section~\ref{sec:experient2}).
The full experimental code for reproducing our results is available in the \texttt{study} directory of the GitHub repository.\footnote{\repourl}

\subsection{Configuration}
\label{app:training_config}


All fine-tuning runs use the same base configuration unless otherwise stated.
We train for three epochs with learning rate $1\mathrm{e}{-5}$, random seeds 42 and 67, per-device batch size 2, gradient accumulation over 2 steps, maximum sequence length 32{,}768, bf16 precision, AdamW optimisation, weight decay 0.01, and a cosine learning rate schedule with warm-up ratio 0.1.
The LoRA rank is $r=16$, with $\alpha=16$ and dropout 0.
We apply LoRA to the standard attention and MLP projection layers used in transformer fine-tuning.
Unless otherwise stated, reported results are averaged across both seeds.

\subsection{Model Details}
\label{app:model_details}

We use four publicly available open-weight reasoning models, all accessed via Hugging Face.
\qwen\footnote{\url{https://huggingface.co/Qwen/Qwen3-8B}}~\citep{QwenQwen38BHugging2025} and \olmo\footnote{\url{https://huggingface.co/allenai/Olmo-3-7B-Think}}~\citep{AllenaiOlmo37BThinkHugging2026} are released under the Apache 2.0 licence; \llama\footnote{\url{https://huggingface.co/deepseek-ai/DeepSeek-R1-Distill-Llama-8B}}~\citep{deepseek-aiDeepSeekR1IncentivizingReasoning2025} is released under the MIT licence; and \nemotron\footnote{\url{https://huggingface.co/nvidia/OpenReasoning-Nemotron-7B}} is released under the Creative Commons Attribution 4.0 licence.
We use the publicly available checkpoints without modifying or redistributing the base model weights.

\subsection{Dataset Details}
\label{app:dataset_details}

This study uses only publicly available dataset sources.
For fine-tuning and in-domain evaluation, we use the Chemistry L-3 subset of \texttt{SciKnowEval}~\citep{fengSciKnowEvalComprehensiveDataset2025}, available under the MIT license on Hugging Face.\footnote{\url{https://huggingface.co/datasets/hicai-zju/SciKnowEval}}
Specifically, we use the processed version from prior work~\citep{shenfeldSelfDistillationEnablesContinual2026}, which augments the original questions with answer explanations, and releases the updated dataset publicly on GitHub.\footnote{\url{https://github.com/idanshen/Self-Distillation}}
The \chemistry\ subset contains 2,674 training examples and 507 held-out evaluation examples.
For mathematical reasoning evaluation, we use \gsmk~\citep{cobbeTrainingVerifiersSolve2021b}, available on Hugging Face under the MIT license.\footnote{\url{https://huggingface.co/datasets/openai/gsm8k}}
For code-generation evaluation, we use the \evalplus~\citep{liuYourCodeGenerated2023a} versions of HumanEval\footnote{\url{https://huggingface.co/datasets/evalplus/humanevalplus}} and MBPP,\footnote{\url{https://huggingface.co/datasets/evalplus/mbppplus}} both available on Hugging Face under the Apache License 2.0.
All datasets are used for research evaluation and fine-tuning in accordance with their stated licenses.

\subsection{Evaluation Details}
\label{app:evaluation_details}

All evaluations use \texttt{vLLM==0.7.3} with \texttt{transformers==4.57.6} and greedy decoding (\texttt{temperature=0}, \texttt{top\_p=1}, \texttt{top\_k} disabled), generating one response per prompt without a system prompt.
We use a minimum generation length of 32 tokens and set \texttt{max\_tokens} to 32,768.
A shared context limit of 32,768 tokens applies to all models, so the effective generation allowance is 32,768 tokens minus the prompt length.
Generation stops when the model produces an end-of-sequence token or reaches this limit.
We record incomplete reasoning traces that reach the limit as overflow, distinguishing them from traces that terminate early without closing the reasoning block.

For each checkpoint, we evaluate on a fixed subset of 256 examples per evaluation dataset, sampled once with random seed 42 and reused across all runs.
For \gsmk, we sample from the official test set.
For \chemistry, we sample from the held-out Chemistry L-3 evaluation split.
For \evalplus, we sample uniformly without stratification from the combined pool of 542 problems, comprising 164 \texttt{HumanEval} and 378 \texttt{MBPP} problems.
The resulting 256-example subset contains 83 \texttt{HumanEval} and 173 \texttt{MBPP} problems.
Because the same fixed subset and checker are used across all checkpoints and strategies, we use these scores for within-study comparisons of training dynamics rather than as claims about full-benchmark performance.

For \gsmk\ and \chemistry, prompts are augmented with the instruction:
\begin{quote}
\small
\texttt{Please provide your final answer in a \textbackslash boxed\{\} environment.}
\end{quote}
We use permissive answer checking for these tasks, allowing correct answers to be recognised across common equivalent formats rather than requiring the answer to appear only in the requested \texttt{\textbackslash boxed\{\}} form.
For \evalplus, prompts are constructed from the benchmark task descriptions.
\texttt{HumanEval} tasks use the prefix:
\begin{quote}
\small
\texttt{Complete the following Python function:\textbackslash n\textbackslash n}
\end{quote}
\texttt{MBPP} tasks additionally specify the required function name:
\begin{quote}
\small
\texttt{\textbackslash n\textbackslash nYour function should be named `\{entry\_point\}'.}
\end{quote}
Generated code is evaluated using the standard \evalplus\ test harness for the corresponding \texttt{HumanEval} or \texttt{MBPP} task.

\subsection{Formatting Examples}
\label{app:formatting_examples}

The examples below illustrate the logical structure of each format, not the exact token sequence.
In practice, \thinkpack\ renders each example using the target model's chat template.
Our training data contains instruction--response pairs, but no explicit reasoning traces.
Experiment 1 (Section~\ref{sec:experient1}) compares two ways of representing this missing reasoning.

In the \textit{empty-think} format, the target includes an empty reasoning block before the final answer:

\begin{quote}
\small
\texttt{<user>}\\
\texttt{\{instruction\}}\\
\texttt{</user>}\\
\texttt{<assistant>}\\
\texttt{<think>}\\
\texttt{</think>}\\
\texttt{\{response\}}\\
\texttt{</assistant>}
\end{quote}

In the \textit{no-think} format, the target contains only the final answer:

\begin{quote}
\small
\texttt{<user>}\\
\texttt{\{instruction\}}\\
\texttt{</user>}\\
\texttt{<assistant>}\\
\texttt{\{response\}}\\
\texttt{</assistant>}
\end{quote}

The key difference is whether the absence of reasoning is represented explicitly through an empty reasoning block, or implicitly by omitting the reasoning segment entirely.

\subsection{Masking Examples}
\label{app:masking_examples}

The mitigation experiments use the \textit{empty <think>} format, but change which tokens contribute to the training loss.
In the examples below, the regions highlighted \masked{red} are masked and therefore excluded from the loss, regions highlighted \open{green} are unchanged.

In the \textit{masked-think} setting, the missing reasoning region is excluded from the loss while the final answer remains supervised:

\begin{quote}
\small
\open{\texttt{<user>}}\\
\open{\texttt{\{instruction\}}}\\
\open{\texttt{</user>}}\\
\open{\texttt{<assistant>}}\\
\masked{\texttt{<think>}}\\
\masked{\texttt{</think>}}\\
\open{\texttt{\{response\}}}\\
\open{\texttt{</assistant>}}
\end{quote}

This prevents the model from being rewarded for producing an empty reasoning trace, while still training it to produce the target answer.

In the \textit{response-only} setting, both the prompt and missing reasoning region are masked, so only the final answer contributes to the loss:

\begin{quote}
\small
\masked{\texttt{<user>}}\\
\masked{\texttt{\{instruction\}}}\\
\masked{\texttt{</user>}}\\
\masked{\texttt{<assistant>}}\\
\masked{\texttt{<think>}}\\
\masked{\texttt{</think>}}\\
\open{\texttt{\{response\}}}\\
\open{\texttt{</assistant>}}
\end{quote}

This corresponds to standard response-only supervised fine-tuning, but applied in a reasoning-aware format.
It allows the model to condition on the full prompt and template structure while avoiding a loss signal that encourages empty reasoning.

\subsection{Distillation Details}
\label{app:distillation_details}

For the distillation baseline, we augment the original instruction--response training data with teacher-generated reasoning traces.
We use OpenAI's \texttt{GPT-5-mini} as the teacher model, as the task is narrowly specified and depends on following a precise prompt rather than open-ended generation, a setting for which the model is well suited~\citep{openaiGPT5MiniAPI2025}.
We use \texttt{gpt-5-mini-2025-08-07} through the OpenAI Chat Completions API, providing one user message without a system prompt.
We set \texttt{max\_completion\_tokens} to 4,096, including the model's internal reasoning tokens, and leave the remaining generation parameters at their API defaults.

For each training example, the teacher is given the original instruction and target response, and asked to produce a concise reasoning summary explaining how to arrive at the answer.
The distillation prompt is:

\begin{quote}
\small
\begin{verbatim}
I need assistance constructing a reasoning dataset.
Given the following question and its correct answer, 
give a concise summary of the reasoning steps that 
explain how to arrive at the answer.

Question: {instruction}

Answer: {response}

In your response, give the reasoning steps inside 
<reasoning_steps> tags, and start your reasoning 
with 'Okay, ', for example:
<reasoning_steps>
Okay, [your reasoning steps here]
</reasoning_steps>
\end{verbatim}
\end{quote}

We allow up to two attempts to extract a valid, non-empty reasoning trace for each example.
The first attempt yields 1,915 valid traces, and the second recovers a further 493, giving 2,408 of 2,674 examples (90.1\%).
The remaining 266 examples are retained without a reasoning trace and formatted according to each model's default missing-reasoning strategy.
The training set therefore remains the same size across all strategies.

\subsection{Compute Details}
\label{app:compute}

All training and evaluation runs were performed on an internal cluster using a single NVIDIA H200 GPU per run, with a 96GB Sapphire Rapids CPU host.
Each model--training-setting run, including periodic evaluation every 100 training steps, took approximately one day to complete, and was run on two different random seeds.
The main experiments therefore required approximately 40 GPU-days in total.
Additional runs for the learning-rate sweep, and ablations, took approximately 21 GPU-days.
Preliminary, debugging, and failed runs used a further approximately 6 GPU-days on the same hardware.
Overall, the project used approximately 67 NVIDIA H200 GPU-days.

\begin{figure*}
    \centering
    \includegraphics[width=\textwidth]{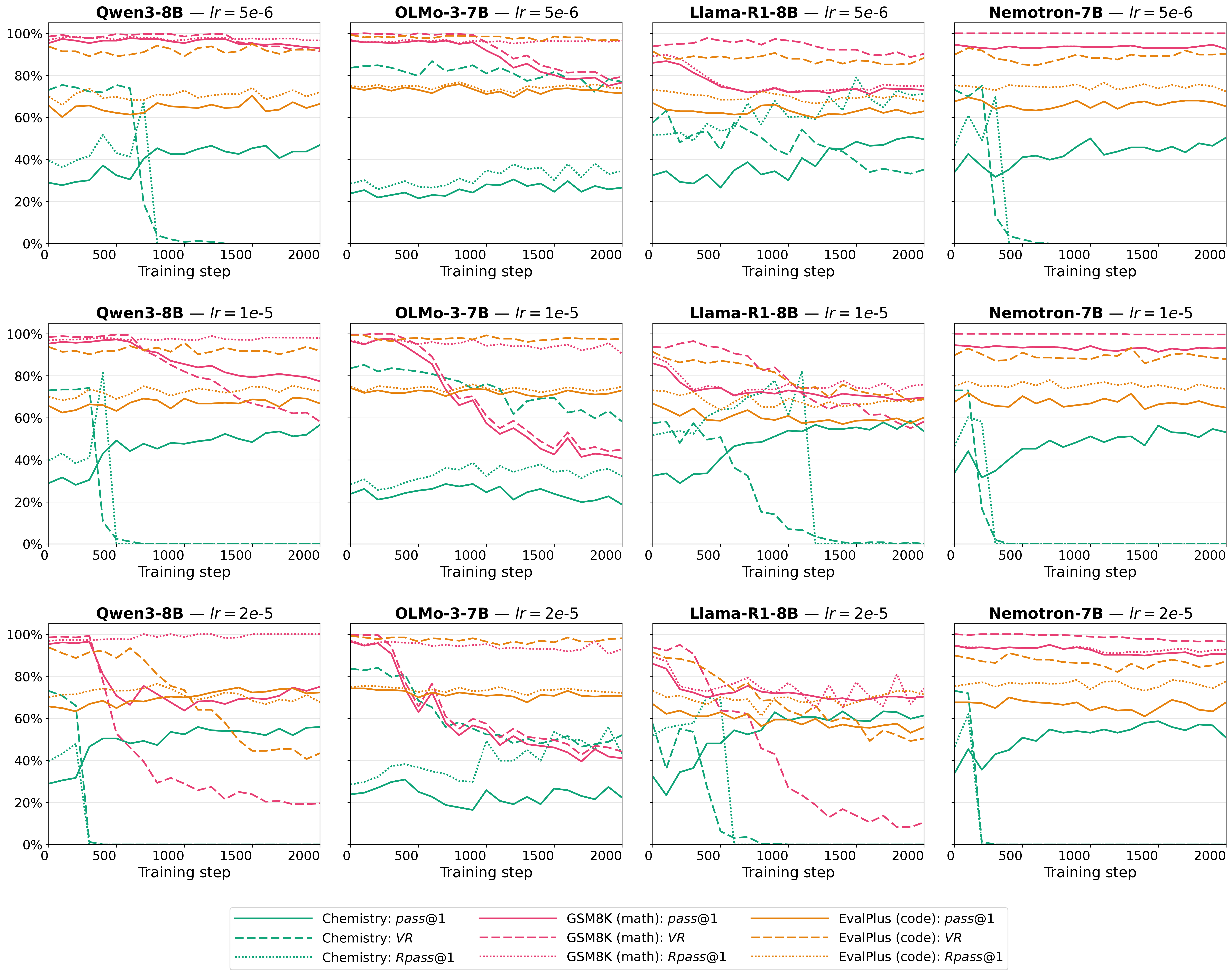}
    \caption{
        \textbf{\textit{Learning-Rate Sensitivity.}}
        We compare three different initial learning rates for standard supervised fine-tuning at $5\mathrm{e}{-6}$, $1\mathrm{e}{-5}$, and $2\mathrm{e}{-5}$.
        Metrics are measured every 100 training steps across three datasets: \chemistry, \gsmk\ (math), and \evalplus\ (code).
        Solid lines show \pass, dashed lines show valid reasoning rate (\vr), and dotted lines show reasoning-conditioned \pass\ (\rpass).
    }
    \label{fig:sweep}
\end{figure*}


\section{Learning-Rate Sensitivity}
\label{app:lr_sweep}

We additionally evaluate whether reasoning-trace collapse is an artefact of a single optimisation setting by sweeping learning rates for the standard fine-tuning setup.
For each model, we perform one fine-tuning run per learning rate, using a single random seed, with learning rates of $5\mathrm{e}{-6}$, $1\mathrm{e}{-5}$, and $2\mathrm{e}{-5}$, while keeping the remaining training configuration fixed.
This sweep is used both to select a stable configuration for the main experiments and to assess how optimisation strength affects the emergence of reasoning-trace collapse.
We use each model's default templating when no reasoning is provided: \textit{empty-think} for \qwen, and \textit{no-think} for \olmo, \llama, and \nemotron.

Figure~\ref{fig:sweep} shows that reasoning-trace collapse is sensitive to learning rate, but not caused by one particular setting.
At lower learning rates, valid reasoning is generally preserved for longer, but in-domain task adaptation is weaker.
At higher learning rates, the model adapts more aggressively, but the loss of valid reasoning becomes faster and more pronounced.


This pattern is clearest for \qwen\, where valid reasoning on \chemistry\ collapses rapidly across settings, and higher learning rates also induce stronger degradation in \gsmk\ and \evalplus\ valid reasoning.
\nemotron\ sees similarly rapid reasoning-trace collapse on \chemistry, but maintains valid reasoning for other datasets throughout for each learning rate.
For \llama, the point at which \chemistry\ reasoning collapses varies with learning rate: lower learning rates delay collapse, while higher learning rates make reasoning behaviour on the other datasets more erratic.
For \olmo, lower learning rates preserve valid reasoning more effectively, while higher learning rates accelerate the drop in \gsmk\ \pass\ and valid reasoning.

Overall, the sweep supports the main finding that reasoning-trace collapse is a robust fine-tuning phenomenon rather than a single hyper-parameter artefact.
Learning rate changes the speed and severity of collapse, but does not remove the underlying trade-off: stronger adaptation can improve or preserve final-answer performance while placing greater pressure on explicit reasoning behaviour.



\section{Additional Training Ablations}
\label{app:training_ablations}

We include two additional ablations to test whether our findings are sensitive to model scale or the choice of fine-tuning dataset.
First, we repeat the main adaptation experiment with \qwenbig, a larger model from the same family as \qwen.
Second, we fine-tune \qwen\ on math and code datasets, rather than \chemistry, to test whether reasoning collapse is specific to the science question-answering setting used in the main experiments.
These ablations are run with a single seed and are intended as targeted robustness checks rather than full replications of the main two-seed experiments.

\subsection{Scaling to a larger model}
\label{app:larger_model}

\begin{figure*}
\centering
\includegraphics[width=\textwidth]{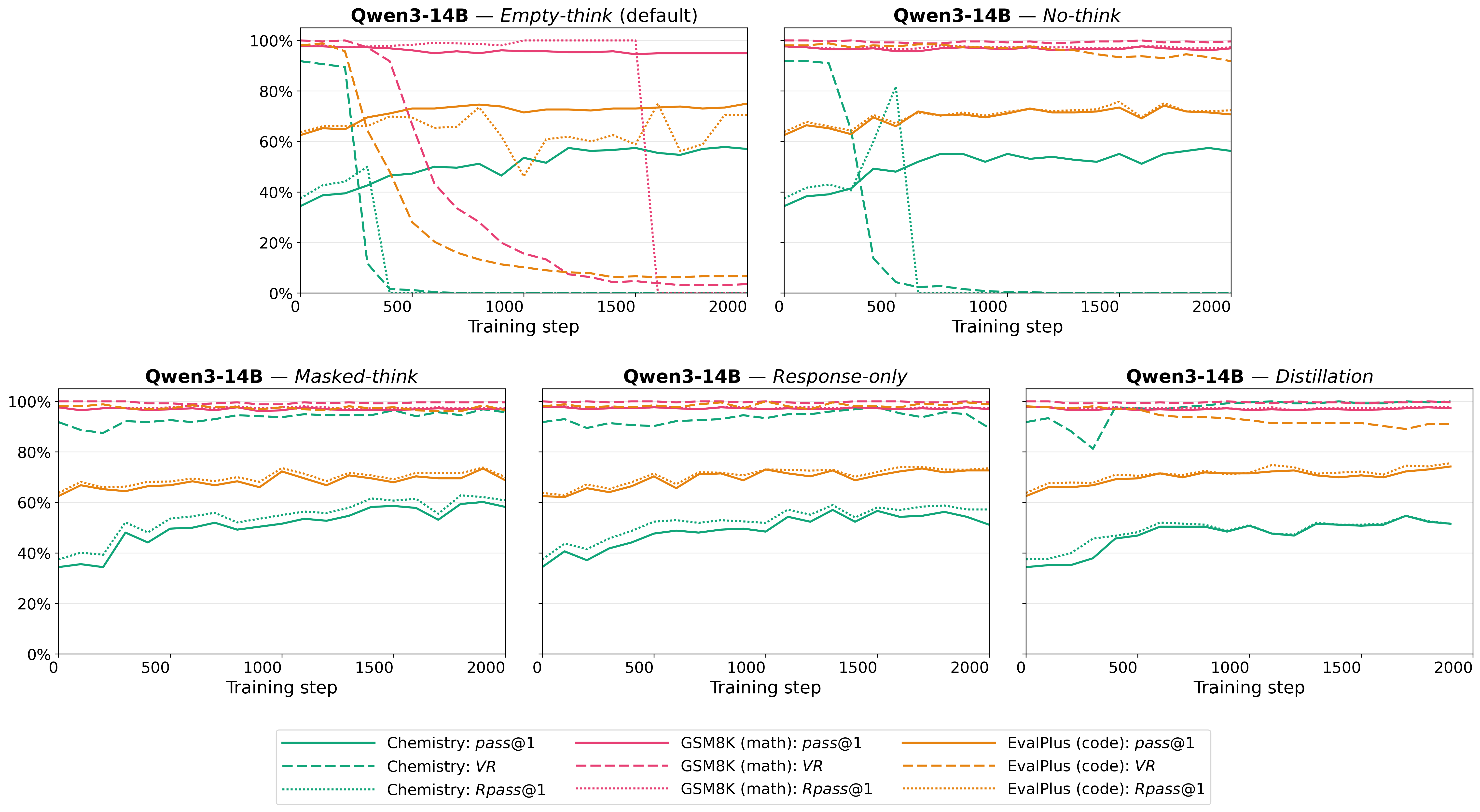}
\caption{
\textbf{\textit{Scaling to a larger model.}}
We compare all fine-tuning strategies for adapting \qwenbig\ to the \chemistry\ dataset.
Metrics are measured every 100 training steps across three evaluation settings: \chemistry, \gsmk\ (math), and \evalplus\ (code).
Solid lines show \pass, dashed lines show valid reasoning rate (\vr), and dotted lines show reasoning-conditioned \pass\ (\rpass).
}
\label{fig:ablation_qwen14b}
\end{figure*}


Figure~\ref{fig:ablation_qwen14b} shows that reasoning collapse is not limited to models in the 7--8B parameter range.
Under \textit{empty-think} and \textit{no-think} training, \qwenbig\ rapidly loses valid reasoning on the \chemistry\ evaluation, despite improving in-domain \pass.
This mirrors the core behaviour observed in the main experiments: fine-tuning can improve adaptation while suppressing the model's ability to produce valid reasoning traces.

The mitigation strategies remain effective at the larger scale.
\textit{Masked-think}, \textit{response-only}, and \textit{distillation} preserve high \vr\ across evaluation settings, while also improving \chemistry\ performance.
This suggests that the benefits of reasoning-aware fine-tuning are not specific to the 7--8B setting.

\subsection{Other Fine-Tuning Datasets}
\label{app:other_datasets}

\begin{figure*}
\centering
\includegraphics[width=\textwidth]{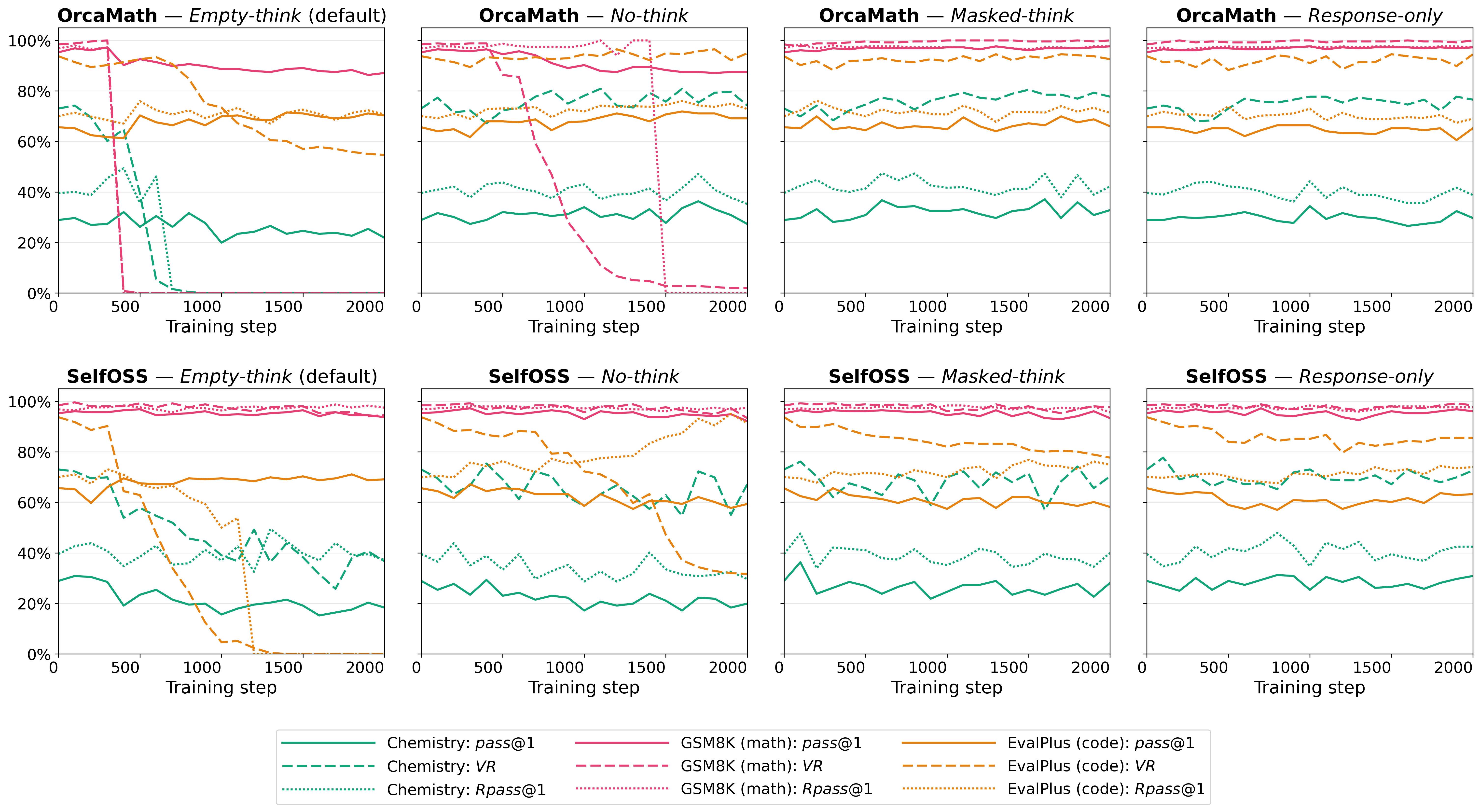}
\caption{
\textbf{\textit{Other fine-tuning datasets.}}
We compare the non-distillation fine-tuning strategies for adapting \qwen\ to \orcamath\ (math) and \selfoss\ (code).
The \texttt{OrcaMath} and \texttt{SelfOSS} fine-tuning subsets are sampled to match the size of the \chemistry\ training set, therefore each contain 2,674 examples.
Metrics are measured every 100 training steps across three evaluation settings: \chemistry, \gsmk\ (math), and \evalplus\ (code).
Solid lines show \pass, dashed lines show valid reasoning rate (\vr), and dotted lines show reasoning-conditioned \pass\ (\rpass).
}
\label{fig:ablation_qwen8b_other_datasets}
\end{figure*}

Figure~\ref{fig:ablation_qwen8b_other_datasets} tests whether reasoning collapse is specific to fine-tuning on \chemistry.
We fine-tune \qwen\ on \orcamath\footnote{\url{https://huggingface.co/datasets/microsoft/orca-math-word-problems-200k}} and \selfoss,\footnote{\url{https://huggingface.co/datasets/bigcode/self-oss-instruct-sc2-exec-filter-50k}} representing math and code adaptation respectively.
Both datasets are sampled to be the same size as the \chemistry\ dataset used for fine-tuning in our other experiments.
We omit distillation in this ablation because generating reasoning traces for each additional dataset substantially increases cost, and because distillation for math and code reasoning has already been widely studied.

Across both datasets, naive \textit{empty-think} training causes severe reasoning collapse, but the pattern differs by evaluation domain.
For \orcamath, valid reasoning rapidly disappears on the in-domain \gsmk\ evaluation, and also collapses on the out-of-domain \chemistry\ evaluation.
In contrast, \evalplus\ retains much more valid reasoning, suggesting that collapse is strongest in domains most aligned with the fine-tuning data.
For \selfoss, the clearest collapse occurs on the in-domain \evalplus\ evaluation, while reasoning on \chemistry\ and \gsmk\ is less uniformly affected.

The mitigation strategies are more stable across both in-domain and out-of-domain evaluations.
\textit{Masked-think} and \textit{response-only} preserve high \vr\ across most settings while maintaining or improving task performance.
\textit{No-think} is less reliable: it can delay or localise collapse in some cases, but does not consistently preserve reasoning across domains.
Overall, these ablations suggest that reasoning collapse is not specific to \chemistry, but is often strongest in the domain targeted by fine-tuning, reinforcing the need to evaluate both in-domain adaptation and out-of-domain reasoning retention.


\section{Additional Robustness and Diagnostic Analyses}
\label{app:additional-analyses}

We provide further analyses of training-seed stability, evaluation-subset robustness, reasoning quality, truncation mechanisms, and the stability of surviving valid reasoning.

\subsection{Training-Seed Stability}
\label{app:seed-stability}

We additionally examine whether the observed training trajectories are consistent across random seeds.
For \qwen\ and \olmo, we compare the two training seeds across the four non-distillation strategies and three evaluation datasets, giving 24 model--strategy--dataset combinations.

The results are highly consistent.
Across these combinations, the mean absolute difference in \vr\ over training never exceeds 3.1 percentage points, while final \vr\ differs by at most 6.6 points.
For \qwen\ under its default empty-think format, Chemistry \vr\ falls to approximately 10\% at step 400 and reaches 0\% by step 700 under both seeds.
\olmo\ similarly follows closely aligned trajectories across all three evaluation datasets.

Both seeds reproduce the main findings: standard fine-tuning can rapidly suppress valid reasoning, while masking substantially improves reasoning retention.
Smaller differences between the two masking strategies remain model- and task-dependent.

\subsection{Evaluation-Subset Robustness}

We also test whether the results depend on the fixed evaluation subsets.
We re-evaluate \qwen\ using independently sampled 256-example subsets from \gsmk\ and \evalplus.

On \gsmk, \vr\ falls by 38.3 percentage points on the original subset and 34.0 points on the new subset.
\evalplus\ remains comparatively stable, with decreases of 2.0 and 4.7 points, respectively.

The same cross-task pattern therefore appears on both subsets: reasoning-trace collapse is substantially stronger on \gsmk\ than on \evalplus.
This suggests that the observed differences are not simply an artefact of the examples originally sampled.

\subsection{Reasoning-Trace Quality}
\label{app:trace-quality}

Our framework deliberately focuses on structural reasoning validity, but the quality of the surviving traces is also important.

We use \texttt{GPT-5-mini} to audit 100 valid \qwen\ \gsmk\ reasoning traces from each of the base and final \chemistry\ fine-tuning checkpoints.
Following a condensed adaptation of \texttt{SelfCheck}~\cite{miaoSelfCheckUsingLLMs2023}, each reasoning step is evaluated using only the problem and preceding steps, without revealing the stated final answer.
This reduces the risk of the judge rationalising its assessment towards the answer.

At the base checkpoint, 94\% of traces are judged correct, 2\% incorrect, and 4\% uncertain.
At the final checkpoint, the corresponding figures are 92\%, 5\%, and 3\%.

Although the quantity of valid reasoning falls substantially, we find no comparable decline in the judged correctness of the traces that remain.
This small audit suggests that the surviving traces generally remain substantively correct, rather than merely satisfying the required output format.

However, this analysis is limited to a sample of one model and evaluation task, and relies on model-assisted judgements.
It does not establish that the reasoning traces are causally responsible for the final answers.

\subsection{Truncation Mechanisms}
\label{app:olmo-analysis}

We further examine the truncation-dominated failures observed for \olmo.
Truncation can occur in two ways: overflow, where the model reaches its generation allowance, or early stopping, where it produces an end-of-sequence token before closing its reasoning block.
Both produce the same structural outcome: an incomplete reasoning trace without a separable final answer.

Table~\ref{tab:truncation-mechanisms} separates these mechanisms for three training strategies.
Under no-think fine-tuning, almost all truncated \chemistry\ responses are full-length overflows: \tr\ is 41.8\%, while overflow accounts for 39.8\%.
On \gsmk, the opposite pattern appears.
\tr\ reaches 55.1\%, but overflow accounts for only 0.4\%, meaning that almost all truncated responses stop early.

\begin{table}[t]
\caption{
    \textbf{\textit{Truncation mechanisms.}}
    \olmo\ truncated reasoning rate (\tr) and overflow rate at the final checkpoint, using seed 42.
    Overflow denotes responses that reach the generation allowance without completing the reasoning structure.
    All values are percentages.
}
\label{tab:truncation-mechanisms}
\centering
\small
\renewcommand{\arraystretch}{1.2}
\setlength{\tabcolsep}{5pt}

\begin{tabular}{lrrrrrr}
\toprule
& \multicolumn{2}{c}{No-think}
& \multicolumn{2}{c}{Empty-think}
& \multicolumn{2}{c}{Distillation} \\
\cmidrule(lr){2-3}
\cmidrule(lr){4-5}
\cmidrule(lr){6-7}
Dataset
& TR & Overflow
& TR & Overflow
& TR & Overflow \\
\midrule
\chemistry\
& 41.8 & 39.8
& 39.1 & 37.1
& 77.7 & 12.1 \\
\gsmk\
& 55.1 & 0.4
& 19.9 & 0.0
& 78.9 & 0.0 \\
\evalplus\
& 2.3 & 2.3
& 2.7 & 2.7
& 2.0 & 2.0 \\
\bottomrule
\end{tabular}
\end{table}

Distillation changes this failure profile.
On \chemistry, \tr\ increases to 77.7\%, but overflow accounts for only 12.1\%.
On \gsmk, \tr\ reaches 78.9\%, while no responses reach the generation limit.
These results show that \olmo's truncation-dominated collapse cannot be explained solely by excessively long reasoning.
Because overflow is defined relative to our shared 32,768-token context limit, a larger budget might allow some overflowed traces to finish, whereas responses that terminate early with an end-of-sequence token would remain incomplete.

This occurs despite the teacher-generated traces containing explicit closing delimiters, and despite the same distillation pipeline preserving high \vr\ for the other models.
One possible explanation is an interaction between \olmo's reinforcement-learning-based post-training and subsequent supervised fine-tuning, although our experiments do not establish this mechanism.

Importantly, we do not claim that the partial reasoning is necessarily incorrect or incoherent.
\tr\ records a structural failure: the model does not complete its reasoning or produce a usable final answer.
Recovering reasoning from incomplete traces would require an additional intervention, which is outside the scope of this study.

\subsection{Stability of Surviving Reasoning}

Finally, we examine whether high \rpass\ is maintained by a consistent set of examples, or whether the valid-and-correct subset changes substantially during training.

We analyse \qwen\ under its default empty-think format on \gsmk, tracking examples that contain both valid reasoning and a correct answer over the final five checkpoints of the seed-42 run.

As \vr\ falls from 65.2\% to 60.5\%, the number of valid-and-correct examples decreases from 162 to 152, while \rpass\ remains between 96.8\% and 98.1\%.
The subset is also stable: its Jaccard overlap with step 1600 remains between 0.85 and 0.89, while overlap between consecutive checkpoints remains between 0.87 and 0.90.
At the final checkpoint, approximately 147 of the 152 valid-and-correct examples were already present at step 1600.

Stable \rpass\ is therefore not produced by a heavily reshuffling subset.
A consistent core of examples continues to produce valid reasoning and correct answers as overall \vr\ declines.
This does not establish that the traces causally produce the answers, but shows that the remaining performance is relatively stable rather than being driven by changing example composition.


\section{Full Metrics}
\label{app:full_metrics}

Tables~\ref{tab:chemistry_baseline},~\ref{tab:chemistry_strategies_posttrain} and~\ref{tab:chemistry_strategies_peakpass} provide the full reasoning-aware metrics for all models, strategies, and evaluation settings.
Table~\ref{tab:chemistry_baseline} reports the base-model results before fine-tuning.
Table~\ref{tab:chemistry_strategies_posttrain} reports the final evaluated checkpoint (step 2000), while Table~\ref{tab:chemistry_strategies_peakpass} reports the checkpoint with the highest \pass\ on \chemistry\ for each model--strategy pair.
Together, these tables distinguish final reasoning retention from peak in-domain performance.

All values are percentages, and fine-tuned point estimates are averaged over two training seeds.
For Table~\ref{tab:chemistry_baseline}, we report 95\% Wilson confidence intervals over the 256 evaluation examples.
For Tables~\ref{tab:chemistry_strategies_posttrain} and~\ref{tab:chemistry_strategies_peakpass}, we report approximate 95\% uncertainty intervals using a conservative combination of Wilson-score intervals.
We calculate a 97.5\% Wilson interval separately for each seed and average the corresponding lower and upper bounds.
By the Bonferroni inequality, this construction accounts for dependence between the seeds without requiring paired per-prompt outcomes.
The reported $\pm$ value is the larger distance from the point estimate to either bound, with the resulting interval clipped to 0--100\%.
These intervals capture evaluation-sample uncertainty, not the full variability across training runs.

We suppress \rpass\ when there are 10 or fewer valid-reasoning responses, as the estimate becomes unstable on very small subsets.
These entries are reported as ``---''.
Bold indicates the best value for each metric and dataset, while $\dagger$ marks the model's default strategy when no reasoning is provided.

\bigskip

\begin{table}[h]
  \centering
  \caption{
    \textbf{\textit{Pre-fine-tuning baseline metrics.}}
        Results for the base models before adaptation (step 0).
        Values are percentages with 95\% uncertainty intervals.
        \pass\ is the accuracy; \rpass\ is the reasoning-conditioned accuracy; \vr\ is the valid reasoning rate; \er\ is the empty reasoning rate; \mr\ is the missing reasoning rate; and \tr\ is the truncated reasoning rate.
  }
  \label{tab:chemistry_baseline}
  \begin{adjustbox}{width=0.8\textwidth}
  \bgroup
  \def\arraystretch{1.4}
  \begin{tabular}{llrrrrrr}
    \toprule
    Model & Dataset & \pass & \rpass & \vr & \er & \mr & \tr \\
    \midrule
    \multirow{3}{*}{\qwen} & \chemistry & 28.9$\pm$5.8 & 39.6$\pm$7.2 & 73.0$\pm$5.7 & 0.0$\pm$1.5 & 0.0$\pm$1.5 & 27.0$\pm$5.7 \\
     & \gsmk & 95.3$\pm$3.3 & 96.8$\pm$3.0 & 98.4$\pm$2.4 & 0.0$\pm$1.5 & 0.0$\pm$1.5 & 1.6$\pm$2.4 \\
     & \evalplus & 65.6$\pm$6.0 & 70.0$\pm$6.1 & 93.8$\pm$3.7 & 0.0$\pm$1.5 & 0.0$\pm$1.5 & 6.2$\pm$3.7 \\
    \midrule
    \multirow{3}{*}{\olmo} & \chemistry & 23.8$\pm$5.6 & 28.5$\pm$6.4 & 83.6$\pm$5.0 & 0.0$\pm$1.5 & 0.0$\pm$1.5 & 16.4$\pm$5.0 \\
     & \gsmk & 96.5$\pm$3.0 & 96.9$\pm$2.9 & 99.6$\pm$1.8 & 0.0$\pm$1.5 & 0.0$\pm$1.5 & 0.4$\pm$1.8 \\
     & \evalplus & 74.2$\pm$5.7 & 74.8$\pm$5.7 & 99.2$\pm$2.0 & 0.0$\pm$1.5 & 0.0$\pm$1.5 & 0.8$\pm$2.0 \\
    \midrule
    \multirow{3}{*}{\llama} & \chemistry & 32.4$\pm$6.0 & 51.7$\pm$8.0 & 57.4$\pm$6.1 & 0.0$\pm$1.5 & 42.6$\pm$6.1 & 0.0$\pm$1.5 \\
     & \gsmk & 85.9$\pm$4.8 & 89.2$\pm$4.6 & 93.8$\pm$3.7 & 0.0$\pm$1.5 & 6.2$\pm$3.7 & 0.0$\pm$1.5 \\
     & \evalplus & 66.8$\pm$6.0 & 73.1$\pm$6.0 & 91.4$\pm$4.1 & 0.0$\pm$1.5 & 8.6$\pm$4.1 & 0.0$\pm$1.5 \\
    \midrule
    \multirow{3}{*}{\nemotron} & \chemistry & 34.0$\pm$6.0 & 46.5$\pm$7.1 & 73.0$\pm$5.7 & 0.0$\pm$1.5 & 0.0$\pm$1.5 & 27.0$\pm$5.7 \\
     & \gsmk & 94.5$\pm$3.5 & 94.5$\pm$3.5 & 100.0$\pm$1.5 & 0.0$\pm$1.5 & 0.0$\pm$1.5 & 0.0$\pm$1.5 \\
     & \evalplus & 67.6$\pm$6.0 & 75.2$\pm$6.0 & 89.8$\pm$4.3 & 0.0$\pm$1.5 & 0.0$\pm$1.5 & 10.2$\pm$4.3 \\
    \bottomrule
  \end{tabular}
  \egroup
  \end{adjustbox}
\end{table}

\begin{table}[p!]
  \centering
  \caption{
        \textbf{\textit{Reasoning-aware metrics at the final evaluated checkpoint.}}
        Results at step 2000, averaged over two training seeds.
        Values are percentages with approximate 95\% uncertainty intervals.
        \pass\ is the accuracy; \rpass\ is the reasoning-conditioned accuracy; \vr\ is the valid reasoning rate; \er\ is the empty reasoning rate; \mr\ is the missing reasoning rate; and \tr\ is the truncated reasoning rate.
    }
  \label{tab:chemistry_strategies_posttrain}
  \centering
  \begin{adjustbox}{width=1.0\textwidth}
  \bgroup
  \def\arraystretch{1.0}
  \begin{tabular}{lllrrrrrr}
    \toprule
    Model & Strategy & Dataset & \pass & \rpass & \vr & \er & \mr & \tr \\
    \midrule
    \multirow{15}{*}{\qwen} & \multirow{3}{*}{\textit{No-think}} & \chemistry & 52.3$\pm$7.0 & --- & 0.0$\pm$1.9 & \textbf{0.0$\pm$1.9} & 100.0$\pm$1.9 & \textbf{0.0$\pm$1.9} \\
     &  & \gsmk & 96.3$\pm$3.7 & 97.2$\pm$3.5 & 96.3$\pm$3.7 & \textbf{0.0$\pm$1.9} & 3.1$\pm$3.5 & 0.6$\pm$2.4 \\
     &  & \evalplus & \textbf{70.5$\pm$6.7} & \textbf{75.2$\pm$6.7} & \textbf{93.8$\pm$4.3} & \textbf{0.0$\pm$1.9} & \textbf{0.0$\pm$1.9} & \textbf{6.2$\pm$4.3} \\
    \cmidrule(lr){2-9}
     & \multirow{3}{*}{\textit{Empty-think}$^\dagger$} & \chemistry & \textbf{54.1$\pm$7.0} & --- & 0.0$\pm$1.9 & 100.0$\pm$1.9 & \textbf{0.0$\pm$1.9} & \textbf{0.0$\pm$1.9} \\
     &  & \gsmk & 79.5$\pm$6.2 & \textbf{98.4$\pm$4.2} & 60.2$\pm$7.0 & 21.7$\pm$6.3 & \textbf{0.0$\pm$1.9} & 18.2$\pm$6.0 \\
     &  & \evalplus & 67.4$\pm$6.8 & 73.2$\pm$6.9 & 92.0$\pm$4.7 & 0.4$\pm$2.2 & \textbf{0.0$\pm$1.9} & 7.6$\pm$4.6 \\
    \cmidrule(lr){2-9}
     & \multirow{3}{*}{\textit{Masked-think}} & \chemistry & 42.8$\pm$7.0 & 52.5$\pm$7.7 & 81.4$\pm$6.0 & 0.2$\pm$2.1 & 0.2$\pm$2.1 & 18.2$\pm$6.0 \\
     &  & \gsmk & 96.5$\pm$3.6 & 96.7$\pm$3.5 & \textbf{99.8$\pm$2.1} & \textbf{0.0$\pm$1.9} & \textbf{0.0$\pm$1.9} & \textbf{0.2$\pm$2.1} \\
     &  & \evalplus & 68.8$\pm$6.8 & 73.9$\pm$6.8 & 93.0$\pm$4.5 & \textbf{0.0$\pm$1.9} & \textbf{0.0$\pm$1.9} & 7.0$\pm$4.5 \\
    \cmidrule(lr){2-9}
     & \multirow{3}{*}{\textit{Response-only}} & \chemistry & 45.1$\pm$7.0 & \textbf{54.1$\pm$7.7} & 83.0$\pm$5.9 & \textbf{0.0$\pm$1.9} & 0.8$\pm$2.5 & 16.2$\pm$5.8 \\
     &  & \gsmk & 96.5$\pm$3.6 & 96.9$\pm$3.5 & 99.6$\pm$2.2 & \textbf{0.0$\pm$1.9} & \textbf{0.0$\pm$1.9} & 0.4$\pm$2.2 \\
     &  & \evalplus & 68.4$\pm$6.8 & 74.0$\pm$6.8 & 92.4$\pm$4.6 & \textbf{0.0$\pm$1.9} & \textbf{0.0$\pm$1.9} & 7.6$\pm$4.6 \\
    \cmidrule(lr){2-9}
     & \multirow{3}{*}{\textit{Distillation}} & \chemistry & 50.8$\pm$6.9 & 51.2$\pm$7.0 & \textbf{99.2$\pm$2.5} & \textbf{0.0$\pm$1.9} & \textbf{0.0$\pm$1.9} & 0.8$\pm$2.5 \\
     &  & \gsmk & \textbf{96.7$\pm$3.5} & 97.1$\pm$3.4 & 99.6$\pm$2.2 & \textbf{0.0$\pm$1.9} & \textbf{0.0$\pm$1.9} & 0.4$\pm$2.2 \\
     &  & \evalplus & 68.6$\pm$6.8 & 74.5$\pm$6.8 & 92.0$\pm$4.6 & \textbf{0.0$\pm$1.9} & \textbf{0.0$\pm$1.9} & 8.0$\pm$4.6 \\
    \midrule
    \multirow{15}{*}{\olmo} & \multirow{3}{*}{\textit{No-think}$^\dagger$} & \chemistry & 19.9$\pm$6.1 & 33.6$\pm$9.0 & 59.2$\pm$7.0 & \textbf{0.0$\pm$1.9} & \textbf{0.0$\pm$1.9} & 40.8$\pm$7.0 \\
     &  & \gsmk & 41.4$\pm$7.0 & 92.2$\pm$7.5 & 44.9$\pm$7.0 & \textbf{0.0$\pm$1.9} & \textbf{0.0$\pm$1.9} & 55.1$\pm$7.0 \\
     &  & \evalplus & 71.7$\pm$6.7 & 73.7$\pm$6.7 & 97.3$\pm$3.3 & \textbf{0.0$\pm$1.9} & \textbf{0.0$\pm$1.9} & 2.7$\pm$3.3 \\
    \cmidrule(lr){2-9}
     & \multirow{3}{*}{\textit{Empty-think}} & \chemistry & 22.9$\pm$6.4 & 36.1$\pm$8.9 & 61.5$\pm$7.0 & 1.6$\pm$2.9 & \textbf{0.0$\pm$1.9} & 36.9$\pm$6.9 \\
     &  & \gsmk & \textbf{74.8$\pm$6.5} & 93.9$\pm$4.9 & \textbf{79.7$\pm$6.2} & \textbf{0.0$\pm$1.9} & \textbf{0.0$\pm$1.9} & \textbf{20.3$\pm$6.2} \\
     &  & \evalplus & 71.1$\pm$6.7 & 72.3$\pm$7.1 & 88.3$\pm$5.2 & 9.4$\pm$4.9 & \textbf{0.0$\pm$1.9} & \textbf{2.3$\pm$3.2} \\
    \cmidrule(lr){2-9}
     & \multirow{3}{*}{\textit{Masked-think}} & \chemistry & 20.7$\pm$6.2 & 33.5$\pm$8.8 & 61.9$\pm$7.0 & \textbf{0.0$\pm$1.9} & \textbf{0.0$\pm$1.9} & 38.1$\pm$7.0 \\
     &  & \gsmk & 64.8$\pm$6.9 & 93.5$\pm$5.4 & 69.3$\pm$6.8 & \textbf{0.0$\pm$1.9} & \textbf{0.0$\pm$1.9} & 30.7$\pm$6.8 \\
     &  & \evalplus & 71.7$\pm$6.7 & 74.0$\pm$6.7 & 96.9$\pm$3.4 & \textbf{0.0$\pm$1.9} & \textbf{0.0$\pm$1.9} & 3.1$\pm$3.4 \\
    \cmidrule(lr){2-9}
     & \multirow{3}{*}{\textit{Response-only}} & \chemistry & \textbf{31.8$\pm$6.8} & 44.7$\pm$8.3 & \textbf{71.3$\pm$6.7} & \textbf{0.0$\pm$1.9} & \textbf{0.0$\pm$1.9} & \textbf{28.7$\pm$6.7} \\
     &  & \gsmk & 49.8$\pm$6.9 & 94.8$\pm$6.1 & 52.5$\pm$7.0 & \textbf{0.0$\pm$1.9} & \textbf{0.0$\pm$1.9} & 47.5$\pm$7.0 \\
     &  & \evalplus & \textbf{73.4$\pm$6.6} & \textbf{75.2$\pm$6.6} & \textbf{97.7$\pm$3.2} & \textbf{0.0$\pm$1.9} & \textbf{0.0$\pm$1.9} & \textbf{2.3$\pm$3.2} \\
    \cmidrule(lr){2-9}
     & \multirow{3}{*}{\textit{Distillation}} & \chemistry & 10.9$\pm$5.1 & \textbf{46.2$\pm$14.1} & 23.6$\pm$6.4 & \textbf{0.0$\pm$1.9} & \textbf{0.0$\pm$1.9} & 76.4$\pm$6.4 \\
     &  & \gsmk & 21.1$\pm$6.2 & \textbf{96.4$\pm$10.4} & 21.9$\pm$6.3 & \textbf{0.0$\pm$1.9} & \textbf{0.0$\pm$1.9} & 78.1$\pm$6.3 \\
     &  & \evalplus & 71.9$\pm$6.7 & 73.6$\pm$6.7 & \textbf{97.7$\pm$3.2} & \textbf{0.0$\pm$1.9} & \textbf{0.0$\pm$1.9} & \textbf{2.3$\pm$3.2} \\
    \midrule
    \multirow{15}{*}{\llama} & \multirow{3}{*}{\textit{No-think}$^\dagger$} & \chemistry & 54.5$\pm$7.0 & --- & 0.2$\pm$2.1 & \textbf{0.0$\pm$1.9} & 99.8$\pm$2.1 & \textbf{0.0$\pm$1.9} \\
     &  & \gsmk & 69.9$\pm$6.8 & 76.0$\pm$8.6 & 58.6$\pm$7.0 & \textbf{0.0$\pm$1.9} & 41.4$\pm$7.0 & \textbf{0.0$\pm$1.9} \\
     &  & \evalplus & 58.4$\pm$7.0 & 67.5$\pm$8.3 & 68.6$\pm$6.8 & \textbf{0.0$\pm$1.9} & 31.4$\pm$6.8 & \textbf{0.0$\pm$1.9} \\
    \cmidrule(lr){2-9}
     & \multirow{3}{*}{\textit{Empty-think}} & \chemistry & 43.2$\pm$7.0 & 45.7$\pm$8.3 & 70.9$\pm$6.7 & 14.5$\pm$5.6 & 14.6$\pm$5.6 & \textbf{0.0$\pm$1.9} \\
     &  & \gsmk & 70.1$\pm$6.7 & 70.1$\pm$6.7 & \textbf{100.0$\pm$1.9} & \textbf{0.0$\pm$1.9} & \textbf{0.0$\pm$1.9} & \textbf{0.0$\pm$1.9} \\
     &  & \evalplus & 60.9$\pm$7.0 & 65.0$\pm$7.2 & 93.2$\pm$4.4 & \textbf{0.0$\pm$1.9} & 6.8$\pm$4.4 & \textbf{0.0$\pm$1.9} \\
    \cmidrule(lr){2-9}
     & \multirow{3}{*}{\textit{Masked-think}} & \chemistry & 41.8$\pm$7.0 & 62.3$\pm$8.9 & 61.7$\pm$7.0 & 0.2$\pm$2.1 & 37.1$\pm$7.0 & 1.0$\pm$2.6 \\
     &  & \gsmk & 71.7$\pm$6.7 & 71.7$\pm$6.7 & \textbf{100.0$\pm$1.9} & \textbf{0.0$\pm$1.9} & \textbf{0.0$\pm$1.9} & \textbf{0.0$\pm$1.9} \\
     &  & \evalplus & 64.1$\pm$6.9 & 69.9$\pm$7.1 & 91.6$\pm$4.7 & \textbf{0.0$\pm$1.9} & 8.4$\pm$4.7 & \textbf{0.0$\pm$1.9} \\
    \cmidrule(lr){2-9}
     & \multirow{3}{*}{\textit{Response-only}} & \chemistry & 38.9$\pm$7.0 & \textbf{66.9$\pm$9.5} & 53.7$\pm$6.9 & 1.2$\pm$2.7 & 44.9$\pm$6.9 & 0.2$\pm$2.1 \\
     &  & \gsmk & \textbf{77.5$\pm$6.3} & \textbf{78.1$\pm$6.3} & 99.0$\pm$2.6 & \textbf{0.0$\pm$1.9} & 1.0$\pm$2.6 & \textbf{0.0$\pm$1.9} \\
     &  & \evalplus & 63.9$\pm$6.9 & \textbf{70.8$\pm$7.1} & 90.2$\pm$5.0 & \textbf{0.0$\pm$1.9} & 9.8$\pm$5.0 & \textbf{0.0$\pm$1.9} \\
    \cmidrule(lr){2-9}
     & \multirow{3}{*}{\textit{Distillation}} & \chemistry & \textbf{55.3$\pm$7.0} & 56.3$\pm$7.1 & \textbf{98.2$\pm$3.0} & \textbf{0.0$\pm$1.9} & \textbf{1.8$\pm$3.0} & \textbf{0.0$\pm$1.9} \\
     &  & \gsmk & 68.4$\pm$6.8 & 68.4$\pm$6.8 & \textbf{100.0$\pm$1.9} & \textbf{0.0$\pm$1.9} & \textbf{0.0$\pm$1.9} & \textbf{0.0$\pm$1.9} \\
     &  & \evalplus & \textbf{66.2$\pm$6.9} & 68.2$\pm$6.9 & \textbf{97.1$\pm$3.4} & \textbf{0.0$\pm$1.9} & \textbf{2.9$\pm$3.4} & \textbf{0.0$\pm$1.9} \\
    \midrule
    \multirow{15}{*}{\nemotron} & \multirow{3}{*}{\textit{No-think}$^\dagger$} & \chemistry & \textbf{53.5$\pm$7.0} & --- & 0.0$\pm$1.9 & \textbf{0.0$\pm$1.9} & 100.0$\pm$1.9 & \textbf{0.0$\pm$1.9} \\
     &  & \gsmk & 92.4$\pm$4.6 & 92.5$\pm$4.6 & 99.4$\pm$2.4 & \textbf{0.0$\pm$1.9} & 0.4$\pm$2.2 & 0.2$\pm$2.1 \\
     &  & \evalplus & 66.4$\pm$6.9 & 75.0$\pm$6.9 & 88.5$\pm$5.2 & \textbf{0.0$\pm$1.9} & \textbf{0.0$\pm$1.9} & 11.5$\pm$5.2 \\
    \cmidrule(lr){2-9}
     & \multirow{3}{*}{\textit{Empty-think}} & \chemistry & 51.4$\pm$7.0 & --- & 0.0$\pm$1.9 & 100.0$\pm$1.9 & \textbf{0.0$\pm$1.9} & \textbf{0.0$\pm$1.9} \\
     &  & \gsmk & 93.0$\pm$4.5 & 93.2$\pm$4.4 & 99.8$\pm$2.1 & \textbf{0.0$\pm$1.9} & \textbf{0.0$\pm$1.9} & 0.2$\pm$2.1 \\
     &  & \evalplus & 66.6$\pm$6.9 & 75.9$\pm$6.9 & 87.7$\pm$5.3 & \textbf{0.0$\pm$1.9} & \textbf{0.0$\pm$1.9} & 12.3$\pm$5.3 \\
    \cmidrule(lr){2-9}
     & \multirow{3}{*}{\textit{Masked-think}} & \chemistry & 44.3$\pm$7.0 & 58.8$\pm$8.1 & 75.4$\pm$6.5 & \textbf{0.0$\pm$1.9} & \textbf{0.0$\pm$1.9} & 24.6$\pm$6.5 \\
     &  & \gsmk & \textbf{94.7$\pm$4.1} & \textbf{94.7$\pm$4.1} & \textbf{100.0$\pm$1.9} & \textbf{0.0$\pm$1.9} & \textbf{0.0$\pm$1.9} & \textbf{0.0$\pm$1.9} \\
     &  & \evalplus & 67.6$\pm$6.8 & 75.5$\pm$6.9 & 89.5$\pm$5.1 & \textbf{0.0$\pm$1.9} & \textbf{0.0$\pm$1.9} & 10.5$\pm$5.1 \\
    \cmidrule(lr){2-9}
     & \multirow{3}{*}{\textit{Response-only}} & \chemistry & 50.0$\pm$6.9 & \textbf{61.0$\pm$7.7} & 82.0$\pm$6.0 & \textbf{0.0$\pm$1.9} & \textbf{0.0$\pm$1.9} & 18.0$\pm$6.0 \\
     &  & \gsmk & 92.6$\pm$4.5 & 92.6$\pm$4.5 & \textbf{100.0$\pm$1.9} & \textbf{0.0$\pm$1.9} & \textbf{0.0$\pm$1.9} & \textbf{0.0$\pm$1.9} \\
     &  & \evalplus & \textbf{71.1$\pm$6.7} & \textbf{76.3$\pm$6.7} & \textbf{93.2$\pm$4.4} & \textbf{0.0$\pm$1.9} & \textbf{0.0$\pm$1.9} & \textbf{6.8$\pm$4.4} \\
    \cmidrule(lr){2-9}
     & \multirow{3}{*}{\textit{Distillation}} & \chemistry & 44.7$\pm$7.0 & 45.7$\pm$7.1 & \textbf{97.9$\pm$3.1} & \textbf{0.0$\pm$1.9} & \textbf{0.0$\pm$1.9} & 2.1$\pm$3.1 \\
     &  & \gsmk & 92.8$\pm$4.5 & 92.8$\pm$4.5 & \textbf{100.0$\pm$1.9} & \textbf{0.0$\pm$1.9} & \textbf{0.0$\pm$1.9} & \textbf{0.0$\pm$1.9} \\
     &  & \evalplus & 66.6$\pm$6.9 & 76.1$\pm$6.9 & 87.5$\pm$5.4 & \textbf{0.0$\pm$1.9} & \textbf{0.0$\pm$1.9} & 12.5$\pm$5.4 \\
    \bottomrule
  \end{tabular}
  \egroup
  \end{adjustbox}
\end{table}

\begin{table}[p!]
  \centering
  \caption{
    \textbf{\textit{Reasoning-aware metrics at peak accuracy.}}
    Results at the checkpoint with the highest \chemistry\ \pass\ for each model--strategy pair, averaged over two training seeds.
    Values are percentages with approximate 95\% uncertainty intervals.
        \pass\ is the accuracy; \rpass\ is the reasoning-conditioned accuracy; \vr\ is the valid reasoning rate; \er\ is the empty reasoning rate; \mr\ is the missing reasoning rate; and \tr\ is the truncated reasoning rate.
}
  \label{tab:chemistry_strategies_peakpass}
  \centering
  \begin{adjustbox}{width=1.0\textwidth}
  \bgroup
  \def\arraystretch{1.0}
  \begin{tabular}{lllrrrrrr}
    \toprule
    Model & Strategy & Dataset & \pass & \rpass & \vr & \er & \mr & \tr \\
    \midrule
    \multirow{15}{*}{\qwen} & \multirow{3}{*}{\textit{No-think}} & \chemistry & \textbf{54.9$\pm$7.0} & --- & 0.0$\pm$1.9 & \textbf{0.0$\pm$1.9} & 100.0$\pm$1.9 & \textbf{0.0$\pm$1.9} \\
     &  & \gsmk & \textbf{96.5$\pm$3.6} & 97.0$\pm$3.5 & 96.5$\pm$3.6 & \textbf{0.0$\pm$1.9} & 3.3$\pm$3.5 & \textbf{0.2$\pm$2.1} \\
     &  & \evalplus & 67.0$\pm$6.9 & 74.0$\pm$6.9 & 90.6$\pm$4.8 & \textbf{0.0$\pm$1.9} & \textbf{0.0$\pm$1.9} & 9.4$\pm$4.8 \\
    \cmidrule(lr){2-9}
     & \multirow{3}{*}{\textit{Empty-think}$^\dagger$} & \chemistry & 54.7$\pm$7.0 & --- & 0.0$\pm$1.9 & 100.0$\pm$1.9 & \textbf{0.0$\pm$1.9} & \textbf{0.0$\pm$1.9} \\
     &  & \gsmk & 78.7$\pm$6.3 & \textbf{97.8$\pm$4.4} & 61.9$\pm$6.9 & 19.3$\pm$6.1 & \textbf{0.0$\pm$1.9} & 18.8$\pm$6.0 \\
     &  & \evalplus & 66.8$\pm$6.9 & 73.2$\pm$6.9 & 91.2$\pm$4.8 & 0.2$\pm$2.1 & \textbf{0.0$\pm$1.9} & 8.6$\pm$4.8 \\
    \cmidrule(lr){2-9}
     & \multirow{3}{*}{\textit{Masked-think}} & \chemistry & 44.1$\pm$7.0 & 52.0$\pm$7.5 & 85.0$\pm$5.7 & \textbf{0.0$\pm$1.9} & \textbf{0.0$\pm$1.9} & 15.0$\pm$5.7 \\
     &  & \gsmk & 96.3$\pm$3.7 & 96.7$\pm$3.6 & \textbf{99.6$\pm$2.2} & \textbf{0.0$\pm$1.9} & \textbf{0.0$\pm$1.9} & 0.4$\pm$2.2 \\
     &  & \evalplus & 68.2$\pm$6.8 & 73.2$\pm$6.9 & \textbf{93.2$\pm$4.4} & \textbf{0.0$\pm$1.9} & \textbf{0.0$\pm$1.9} & \textbf{6.8$\pm$4.4} \\
    \cmidrule(lr){2-9}
     & \multirow{3}{*}{\textit{Response-only}} & \chemistry & 46.7$\pm$7.0 & \textbf{57.2$\pm$7.8} & 81.6$\pm$6.0 & \textbf{0.0$\pm$1.9} & 1.0$\pm$2.6 & 17.4$\pm$5.9 \\
     &  & \gsmk & \textbf{96.5$\pm$3.6} & 97.2$\pm$3.4 & 99.2$\pm$2.5 & \textbf{0.0$\pm$1.9} & \textbf{0.0$\pm$1.9} & 0.8$\pm$2.5 \\
     &  & \evalplus & 68.4$\pm$6.8 & 74.9$\pm$6.8 & 91.2$\pm$4.8 & \textbf{0.0$\pm$1.9} & \textbf{0.0$\pm$1.9} & 8.8$\pm$4.8 \\
    \cmidrule(lr){2-9}
     & \multirow{3}{*}{\textit{Distillation}} & \chemistry & \textbf{54.9$\pm$7.0} & 55.2$\pm$7.0 & \textbf{99.4$\pm$2.4} & \textbf{0.0$\pm$1.9} & \textbf{0.0$\pm$1.9} & 0.6$\pm$2.4 \\
     &  & \gsmk & 96.1$\pm$3.7 & 97.4$\pm$3.3 & 98.6$\pm$2.8 & \textbf{0.0$\pm$1.9} & \textbf{0.0$\pm$1.9} & 1.4$\pm$2.8 \\
     &  & \evalplus & \textbf{70.7$\pm$6.7} & \textbf{75.9$\pm$6.7} & \textbf{93.2$\pm$4.4} & \textbf{0.0$\pm$1.9} & \textbf{0.0$\pm$1.9} & \textbf{6.8$\pm$4.4} \\
    \midrule
    \multirow{15}{*}{\olmo} & \multirow{3}{*}{\textit{No-think}$^\dagger$} & \chemistry & 29.3$\pm$6.7 & 36.9$\pm$7.8 & \textbf{79.3$\pm$6.2} & \textbf{0.0$\pm$1.9} & \textbf{0.0$\pm$1.9} & \textbf{20.7$\pm$6.2} \\
     &  & \gsmk & 79.9$\pm$6.1 & \textbf{95.7$\pm$4.3} & 83.4$\pm$5.7 & \textbf{0.0$\pm$1.9} & \textbf{0.0$\pm$1.9} & 16.6$\pm$5.7 \\
     &  & \evalplus & 71.5$\pm$6.7 & 72.5$\pm$6.7 & \textbf{98.6$\pm$2.7} & \textbf{0.0$\pm$1.9} & \textbf{0.0$\pm$1.9} & \textbf{1.4$\pm$2.7} \\
    \cmidrule(lr){2-9}
     & \multirow{3}{*}{\textit{Empty-think}} & \chemistry & 32.0$\pm$6.8 & 40.5$\pm$7.9 & 79.1$\pm$6.2 & \textbf{0.0$\pm$1.9} & \textbf{0.0$\pm$1.9} & 20.9$\pm$6.2 \\
     &  & \gsmk & \textbf{87.9$\pm$5.2} & 94.9$\pm$4.2 & \textbf{92.6$\pm$4.4} & \textbf{0.0$\pm$1.9} & \textbf{0.0$\pm$1.9} & \textbf{7.4$\pm$4.4} \\
     &  & \evalplus & 71.1$\pm$6.7 & 72.8$\pm$6.7 & 97.5$\pm$3.2 & 0.2$\pm$2.1 & \textbf{0.0$\pm$1.9} & 2.3$\pm$3.2 \\
    \cmidrule(lr){2-9}
     & \multirow{3}{*}{\textit{Masked-think}} & \chemistry & 31.4$\pm$6.8 & 41.8$\pm$8.1 & 75.2$\pm$6.5 & \textbf{0.0$\pm$1.9} & \textbf{0.0$\pm$1.9} & 24.8$\pm$6.5 \\
     &  & \gsmk & 67.8$\pm$6.8 & 93.8$\pm$5.2 & 72.3$\pm$6.7 & \textbf{0.0$\pm$1.9} & \textbf{0.0$\pm$1.9} & 27.7$\pm$6.7 \\
     &  & \evalplus & 71.1$\pm$6.7 & 73.5$\pm$6.7 & 96.7$\pm$3.5 & \textbf{0.0$\pm$1.9} & \textbf{0.0$\pm$1.9} & 3.3$\pm$3.5 \\
    \cmidrule(lr){2-9}
     & \multirow{3}{*}{\textit{Response-only}} & \chemistry & \textbf{36.5$\pm$6.9} & \textbf{48.6$\pm$8.0} & 75.2$\pm$6.5 & \textbf{0.0$\pm$1.9} & \textbf{0.0$\pm$1.9} & 24.8$\pm$6.5 \\
     &  & \gsmk & 53.9$\pm$7.0 & 93.2$\pm$6.2 & 57.8$\pm$7.0 & \textbf{0.0$\pm$1.9} & \textbf{0.0$\pm$1.9} & 42.2$\pm$7.0 \\
     &  & \evalplus & \textbf{74.0$\pm$6.6} & \textbf{75.5$\pm$6.5} & 98.0$\pm$3.0 & \textbf{0.0$\pm$1.9} & \textbf{0.0$\pm$1.9} & 2.0$\pm$3.0 \\
    \cmidrule(lr){2-9}
     & \multirow{3}{*}{\textit{Distillation}} & \chemistry & 32.0$\pm$6.8 & 41.7$\pm$8.0 & 76.8$\pm$6.4 & \textbf{0.0$\pm$1.9} & \textbf{0.0$\pm$1.9} & 23.2$\pm$6.4 \\
     &  & \gsmk & 60.4$\pm$6.7 & 95.2$\pm$5.5 & 63.3$\pm$6.6 & \textbf{0.0$\pm$1.9} & \textbf{0.0$\pm$1.9} & 36.7$\pm$6.6 \\
     &  & \evalplus & 71.7$\pm$6.7 & 73.7$\pm$6.7 & 97.3$\pm$3.3 & \textbf{0.0$\pm$1.9} & \textbf{0.0$\pm$1.9} & 2.7$\pm$3.3 \\
    \midrule
    \multirow{15}{*}{\llama} & \multirow{3}{*}{\textit{No-think}$^\dagger$} & \chemistry & \textbf{59.2$\pm$7.0} & --- & 1.4$\pm$2.8 & \textbf{0.0$\pm$1.9} & 98.6$\pm$2.8 & \textbf{0.0$\pm$1.9} \\
     &  & \gsmk & 70.7$\pm$6.7 & 75.5$\pm$8.5 & 60.5$\pm$6.9 & \textbf{0.0$\pm$1.9} & 39.5$\pm$6.9 & \textbf{0.0$\pm$1.9} \\
     &  & \evalplus & 58.0$\pm$7.0 & 67.9$\pm$8.2 & 71.1$\pm$6.7 & \textbf{0.0$\pm$1.9} & 28.9$\pm$6.7 & \textbf{0.0$\pm$1.9} \\
    \cmidrule(lr){2-9}
     & \multirow{3}{*}{\textit{Empty-think}} & \chemistry & 47.3$\pm$7.0 & 50.1$\pm$8.2 & 71.7$\pm$6.7 & 14.8$\pm$5.6 & 13.5$\pm$5.5 & \textbf{0.0$\pm$1.9} \\
     &  & \gsmk & 69.7$\pm$6.8 & 69.7$\pm$6.8 & \textbf{100.0$\pm$1.9} & \textbf{0.0$\pm$1.9} & \textbf{0.0$\pm$1.9} & \textbf{0.0$\pm$1.9} \\
     &  & \evalplus & 60.9$\pm$7.0 & 65.5$\pm$7.2 & 92.8$\pm$4.5 & \textbf{0.0$\pm$1.9} & 7.2$\pm$4.5 & \textbf{0.0$\pm$1.9} \\
    \cmidrule(lr){2-9}
     & \multirow{3}{*}{\textit{Masked-think}} & \chemistry & 44.7$\pm$7.0 & 63.8$\pm$8.6 & 65.4$\pm$6.9 & 0.4$\pm$2.2 & 33.8$\pm$6.9 & 0.4$\pm$2.2 \\
     &  & \gsmk & 70.1$\pm$6.7 & 70.1$\pm$6.7 & \textbf{100.0$\pm$1.9} & \textbf{0.0$\pm$1.9} & \textbf{0.0$\pm$1.9} & \textbf{0.0$\pm$1.9} \\
     &  & \evalplus & 60.9$\pm$7.0 & 67.4$\pm$7.2 & 90.4$\pm$4.9 & \textbf{0.0$\pm$1.9} & 9.6$\pm$4.9 & \textbf{0.0$\pm$1.9} \\
    \cmidrule(lr){2-9}
     & \multirow{3}{*}{\textit{Response-only}} & \chemistry & 46.1$\pm$7.0 & \textbf{72.8$\pm$8.8} & 59.0$\pm$7.0 & 1.0$\pm$2.6 & 40.0$\pm$7.0 & \textbf{0.0$\pm$1.9} \\
     &  & \gsmk & \textbf{77.9$\pm$6.3} & \textbf{78.2$\pm$6.3} & 99.6$\pm$2.2 & \textbf{0.0$\pm$1.9} & 0.4$\pm$2.2 & \textbf{0.0$\pm$1.9} \\
     &  & \evalplus & 64.6$\pm$6.9 & \textbf{72.7$\pm$7.1} & 88.9$\pm$5.2 & \textbf{0.0$\pm$1.9} & 11.1$\pm$5.2 & \textbf{0.0$\pm$1.9} \\
    \cmidrule(lr){2-9}
     & \multirow{3}{*}{\textit{Distillation}} & \chemistry & 57.4$\pm$7.0 & 58.2$\pm$7.1 & \textbf{98.2$\pm$3.0} & \textbf{0.0$\pm$1.9} & \textbf{1.8$\pm$3.0} & \textbf{0.0$\pm$1.9} \\
     &  & \gsmk & 69.7$\pm$6.8 & 69.7$\pm$6.8 & \textbf{100.0$\pm$1.9} & \textbf{0.0$\pm$1.9} & \textbf{0.0$\pm$1.9} & \textbf{0.0$\pm$1.9} \\
     &  & \evalplus & \textbf{66.2$\pm$6.9} & 68.5$\pm$6.9 & \textbf{96.7$\pm$3.5} & \textbf{0.0$\pm$1.9} & \textbf{3.3$\pm$3.5} & \textbf{0.0$\pm$1.9} \\
    \midrule
    \multirow{15}{*}{\nemotron} & \multirow{3}{*}{\textit{No-think}$^\dagger$} & \chemistry & 55.3$\pm$7.0 & --- & 0.0$\pm$1.9 & \textbf{0.0$\pm$1.9} & 100.0$\pm$1.9 & \textbf{0.0$\pm$1.9} \\
     &  & \gsmk & 91.6$\pm$4.7 & 91.6$\pm$4.7 & 99.8$\pm$2.1 & \textbf{0.0$\pm$1.9} & 0.2$\pm$2.1 & \textbf{0.0$\pm$1.9} \\
     &  & \evalplus & \textbf{68.4$\pm$6.8} & 76.6$\pm$6.8 & 89.3$\pm$5.1 & \textbf{0.0$\pm$1.9} & \textbf{0.0$\pm$1.9} & 10.7$\pm$5.1 \\
    \cmidrule(lr){2-9}
     & \multirow{3}{*}{\textit{Empty-think}} & \chemistry & 54.9$\pm$7.0 & --- & 0.0$\pm$1.9 & 100.0$\pm$1.9 & \textbf{0.0$\pm$1.9} & \textbf{0.0$\pm$1.9} \\
     &  & \gsmk & \textbf{93.6$\pm$4.3} & \textbf{93.6$\pm$4.3} & \textbf{100.0$\pm$1.9} & \textbf{0.0$\pm$1.9} & \textbf{0.0$\pm$1.9} & \textbf{0.0$\pm$1.9} \\
     &  & \evalplus & 67.8$\pm$6.8 & 75.6$\pm$6.9 & 89.6$\pm$5.1 & \textbf{0.0$\pm$1.9} & \textbf{0.0$\pm$1.9} & 10.4$\pm$5.1 \\
    \cmidrule(lr){2-9}
     & \multirow{3}{*}{\textit{Masked-think}} & \chemistry & 50.6$\pm$6.9 & 65.7$\pm$7.9 & 77.0$\pm$6.4 & \textbf{0.0$\pm$1.9} & \textbf{0.0$\pm$1.9} & 23.0$\pm$6.4 \\
     &  & \gsmk & 93.4$\pm$4.4 & 93.4$\pm$4.4 & \textbf{100.0$\pm$1.9} & \textbf{0.0$\pm$1.9} & \textbf{0.0$\pm$1.9} & \textbf{0.0$\pm$1.9} \\
     &  & \evalplus & \textbf{68.4$\pm$6.8} & 75.3$\pm$6.8 & \textbf{90.8$\pm$4.8} & \textbf{0.0$\pm$1.9} & \textbf{0.0$\pm$1.9} & \textbf{9.2$\pm$4.8} \\
    \cmidrule(lr){2-9}
     & \multirow{3}{*}{\textit{Response-only}} & \chemistry & \textbf{56.1$\pm$7.0} & \textbf{66.9$\pm$7.5} & 83.8$\pm$5.8 & \textbf{0.0$\pm$1.9} & 0.6$\pm$2.3 & 15.6$\pm$5.7 \\
     &  & \gsmk & 92.8$\pm$4.5 & 92.8$\pm$4.5 & \textbf{100.0$\pm$1.9} & \textbf{0.0$\pm$1.9} & \textbf{0.0$\pm$1.9} & \textbf{0.0$\pm$1.9} \\
     &  & \evalplus & \textbf{68.4$\pm$6.8} & 76.4$\pm$6.8 & 89.5$\pm$5.1 & \textbf{0.0$\pm$1.9} & \textbf{0.0$\pm$1.9} & 10.5$\pm$5.1 \\
    \cmidrule(lr){2-9}
     & \multirow{3}{*}{\textit{Distillation}} & \chemistry & 49.2$\pm$7.0 & 49.7$\pm$7.0 & \textbf{99.0$\pm$2.6} & \textbf{0.0$\pm$1.9} & \textbf{0.0$\pm$1.9} & 1.0$\pm$2.6 \\
     &  & \gsmk & 93.4$\pm$4.4 & 93.4$\pm$4.4 & \textbf{100.0$\pm$1.9} & \textbf{0.0$\pm$1.9} & \textbf{0.0$\pm$1.9} & \textbf{0.0$\pm$1.9} \\
     &  & \evalplus & \textbf{68.4$\pm$6.8} & \textbf{76.8$\pm$6.8} & 89.1$\pm$5.1 & \textbf{0.0$\pm$1.9} & \textbf{0.0$\pm$1.9} & 10.9$\pm$5.1 \\
    \bottomrule
  \end{tabular}
  \egroup
  \end{adjustbox}
\end{table}






\end{document}